\documentclass[10pt,twocolumn,letterpaper]{article}

\usepackage[pagenumbers]{cvpr} %

\usepackage{graphicx}
\usepackage{amsmath}
\usepackage{amssymb}
\usepackage{booktabs}
\usepackage{multirow}
\usepackage{xcolor}
\usepackage{arydshln}
\usepackage{wrapfig}
\usepackage{enumitem}
\usepackage{microtype}
\usepackage[font=small]{caption}

\newlength{\imagecolwidth}
\definecolor{newlightblue}{RGB}{0,75,255}
\usepackage[pagebackref,breaklinks,colorlinks, urlcolor={newlightblue}, citecolor={newlightblue}]{hyperref} 

\makeatletter
\def\adl@drawiv#1#2#3{%
        \hskip.5\tabcolsep
        \xleaders#3{#2.5\@tempdimb #1{1}#2.5\@tempdimb}%
                #2\z@ plus1fil minus1fil\relax
        \hskip.5\tabcolsep}
\newcommand{\cdashlinelr}[1]{%
  \noalign{\vskip\aboverulesep
           \global\let\@dashdrawstore\adl@draw
           \global\let\adl@draw\adl@drawiv}
  \cdashline{#1}
  \noalign{\global\let\adl@draw\@dashdrawstore
           \vskip\belowrulesep}}
\makeatother

\usepackage[capitalize]{cleveref}
\crefname{section}{Sec.}{Secs.}
\Crefname{section}{Section}{Sections}
\Crefname{table}{Table}{Tables}
\crefname{table}{Tab.}{Tabs.}

\def\cvprPaperID{4785} %
\def\confName{CVPR}
\def\confYear{2023}

\DeclareMathOperator*{\argmax}{arg\,max}
\newcommand{\bx}[0]{{\mathbf x}}

\newcommand{\mypar}[1]{\vspace{-3mm}\paragraph{#1}}
\def\upvspacefig{\vspace{-0.0mm}}
\newcommand{\supparxiv}[2]{#2}

\begin{document}

\title{Self-Supervised Video Forensics by Audio-Visual Anomaly Detection}

\author{Chao Feng \qquad Ziyang Chen \qquad Andrew Owens \vspace{3.5mm}\\ 
University of Michigan~~~~\\
}
\maketitle

\begin{abstract}
Manipulated videos often contain subtle inconsistencies between their visual and audio signals. We propose a video forensics method, based on anomaly detection, that can identify these inconsistencies, and that can be trained solely using real, unlabeled data. We train an autoregressive model to generate sequences of audio-visual features, using feature sets that capture the temporal synchronization between video frames and sound. At test time, we then flag videos that the model assigns low probability. Despite being trained entirely on real videos, our model obtains strong performance on the task of detecting manipulated speech videos. Project site: \href{https://cfeng16.github.io/audio-visual-forensics/}{https://cfeng16.github.io/audio-visual-forensics}.

\end{abstract}

\section{Introduction}
\label{sec:intro}

Supervised learning underlies today's most successful methods for image and video forensics. However, the difficulty of collecting large, labeled datasets that fully capture all of the possible manipulations that one might encounter in the wild places significant limitations on this approach. A longstanding goal of the forensics community has been to design methods that, instead, learn to detect manipulations using cues discovered by analyzing large amounts of {\em real} data through self-supervision~\cite{huh2018fighting,cozzolino2019noiseprint}.

\looseness=-1
We propose a method that identifies manipulated video through {\em anomaly detection}. Our model learns how audio and visual data temporally co-occur by training on large amounts of real, unlabeled video. At test time, we can then flag videos that our model assigns low probability, such as those whose video and audio streams are inconsistent.

\begin{figure}[t!]
    \centering
    \vspace{-2mm}
    \includegraphics[width=\linewidth]{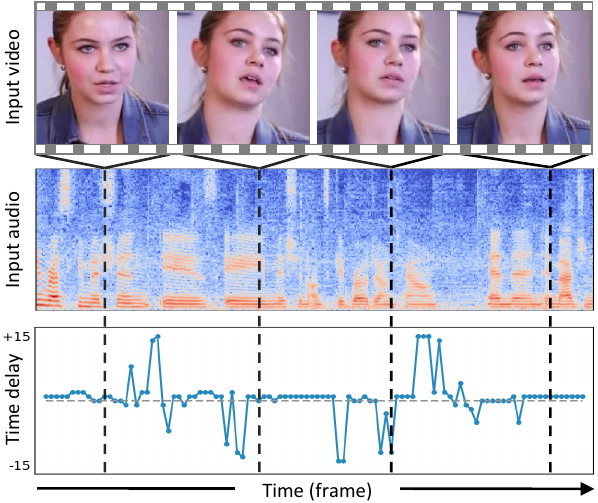}
    
      \vspace{-0.5mm}
            \caption{{\bf Audio-visual anomaly detection.}  We identify fake videos by finding anomalies in their audio-visual features, using generative models trained entirely on {\em real} videos. In one variation of our model (shown here), we use the {\em time delay} between the two modalities as our feature set, i.e., temporal misalignment between each video frame and the audio stream. We learn the distribution of these sequences, then flag sequences with low probability.
      } 
      \vspace{-2mm}
    \label{teaser}
\end{figure}

One might expect that this problem could be posed as simply {detecting} {out-of-sync} examples, such as by finding  cases in which a speaker's mouth does not open precisely at the onset of a spoken word.
Unfortunately, videos in the wild are often ``naturally'' misaligned due to errors in encoding or recording, such as by having a single, consistent shift by a few frames~\cite{chung2016out,afouras2021selfsupervised}.

Instead, we pose the problem as detecting anomalies in what we call {\em synchronization features}: audio-visual features that are designed to convey the temporal alignment between vision and sound. We evaluate several feature sets, each extracted from a model that has been trained to temporally align audio and visual streams of a video~\cite{chen2021audio,owens2018audio,chung2016out}. In Figure~\ref{teaser}, we show one such feature set:  the amount of time that each video frame appears to be temporally offset from its corresponding sound. To detect anomalies, we fit an autoregressive generative model~\cite{vaswani2017attention,radford2019language} to sequences of synchronization features extracted from real videos, and identify low probability examples.

A key advantage of our formulation is that it does not require any manipulated examples for training. It also does not require the speakers in the test set to already be present in the training set. This is in contrast to previous audio-visual forensics approaches, which either require finetuning on datasets of manipulated video~\cite{haliassos2022leveraging}, or which are based on verifying that the speaker's voice matches previously observed examples~\cite{cozzolino2022audio}.

We evaluate our model on videos that have manipulated a person's speech and face, using datasets of lip-synced and audio-driven face reenactment videos, some of which are also manipulated by faceswap techniques. Our model obtains strong performance on the FakeAVCeleb~\cite{NEURIPS_DATASETS_AND_BENCHMARKS2021_d9d4f495} and KoDF~\cite{kwon2021kodf} datasets, despite the fact that it is trained entirely on real examples obtained from other video datasets. Our model generalizes to other spoken languages without retraining and obtains robustness to a variety of postprocessing operations, such as compression and blurring. We show through our experiments that: %
\begin{itemize}[leftmargin=*,topsep=1pt, noitemsep]
   \item Video forensics can be posed as an audio-visual anomaly detection problem.
   \item Synchronization features convey information about video manipulations.
   \item Our model can successfully detect fake videos, while training solely on real videos.
   \item Our model generalizes to many types of image postprocessing operations and to speech videos from spoken languages not observed during training.
\end{itemize}

\section{Related Work}
\label{sec:related}

\paragraph{Audio-visual forensics.}
In early work, Malik and Farid~\cite{malik2010audio} detected audio manipulations by finding inconsistencies in reverberation. Recent work has focused on detecting manipulated speech videos using audio-visual inconsistencies.
Several approaches have directly trained audio-visual networks through supervised learning, using labels indicating whether a video is manipulated~\cite{mittal2020emotions, chugh2020not}. A variety of methods have recently used audio-visual self-supervision for pretraining supervised models, which are finetuned with ``real or fake'' labels. Zeng \etal~\cite{zeng2021contrastive} used local and global contrastive learning methods to learn video features. Haliassos \etal~\cite{haliassos2022leveraging} jointly solved a negative-less contrastive learning problem~\cite{grill2020bootstrap} and a forensics task. Other work~\cite{haliassos2021lips} pretrains using lip-reading data. Zhou and Lim~\cite{zhou2021joint} used audio-visual synchronization signal implicitly, and proposed a dataset for audio-visual deepfake detection\footnote{Their dataset is not publicly available.}. In contrast to these methods, our approach is trained entirely using {\em real} data and does not require any labels or examples of fake videos. Other work has used speaker verification~\cite{cozzolino2022audio} and phoneme-viseme mismatches~\cite{agarwal2020detecting} to detect fake videos and it also detects face swap manipulations, which preserve the synchronization between modalities. In contrast, our approach detects misaligned images and sounds and does not require that examples from the speaker be present in the training set.

\mypar{Audio-visual representation learning.} 
A variety of methods have been proposed to learn audio-visual representation from videos via self-supervision. Researchers have leveraged the natural semantic correspondence in the videos between frames and audio tracks~\cite{asano2020labelling,zeng2021contrastive,morgado2021audio} to learn multi-modal features and applied them to downstream tasks such as sound localization~\cite{hu2022mix,arandjelovic2018objects,mo2022_ezvsl}. Other work studies temporal synchronization between audio and visual signals to learn audio-visual features~\cite{chung2016out,owens2018audio,korbar2018cooperative}, which can be used for active speaker detection~\cite{tao2021someone,alcazar2021maas,kopuklu2021design}, source separation~\cite{majumder2021move2hear,gao2021visualvoice,zhou2020sep}, lip reading~\cite{afouras2020asr,martinez2020lipreading,ma2021towards} and so on. Our method uses the off-the-shelf audio-visual synchronization model to perform anomaly detection.

\mypar{Visual face forensics.} A major focus of the forensics field has been on the problem of detecting manipulated videos of human faces. In recent years, a variety of visual face manipulation datasets are proposed, such as FaceForensics++~\cite{rossler2019faceforensics++}, VideoForensicsHQ~\cite{fox2021videoforensicshq} and \textit{FFIW}$_{\text{10K}}$~\cite{zhou2021face}. Meanwhile, many methods are proposed to detect synthetic contents to fight against their potential threats. Some work~\cite{bianchi2012image, li2018exposing, guo2022eyes} has proposed to use hand-crafted features to capture inconsistent visual or JPEG artifacts. Other work has proposed to use deep learning to inspect specific artifacts, such as blending~\cite{li2020face}, frequency domain~\cite{qian2020thinking, durall2020watch, frank2020leveraging}, or texture~\cite{liu2020global}. A variety of methods have studied the generalization between detection classifiers~\cite{wang2019cnn,chai2020makes}.

\mypar{Anomaly detection.} 
A variety of methods have learned a distribution, then flagged unusual examples. These examples are often considered anomalies~\cite{zong2018deep, schlegl2017unsupervised, zenati2018adversarially,liu2019generative} or outliers~\cite{sabokrou2018adversarially, pidhorskyi2018generative}, and are used as part of open-set recognition~\cite{kong2021opengan, zhang2020hybrid}. 
We formulate video forensics as the task of detecting anomalies, using a feature set that conveys information that would be hard for a forger to create. There have been a variety of methods proposed for learning this distribution, such as  GAN discriminators~\cite{schlegl2017unsupervised,sabokrou2018adversarially,pidhorskyi2018generative,zenati2018adversarially,liu2019generative,kong2021opengan,zong2018deep,NIPS2014_5ca3e9b1}, flow-based models~\cite{zhang2020hybrid}, and autoregressive models~\cite{song2017pixeldefend}. Similarly, our model is based on an autoregressive generative model~\cite{radford2019language,bengio2000neural}, since they have achieved strong performance at modeling complex distributions. Other work addresses goals similar to anomaly detection by creating methods that model uncertainty~\cite{liang2017enhancing} or that leverage outlier exposure~\cite{dhamija2018reducing,ruff2019deep,hendrycks2018deep}. Some work~\cite{huh2018fighting,cozzolino2016single,kalyan2018satellite,d2017autoencoder,bondi2017tampering,khalid2020oc,perez2019deep} has used special-purpose anomaly detection methods for image/video forensics. Other work~\cite{fei2022learning,zhao2021learning,wu2019mantra,hu2021dynamic} uses supervised learning to find anomalous patterns. In contrast, our method builds the likelihood function entirely on real videos and views low-probability examples as fake.

\section{Method}
\begin{figure*}[t!]
    \centering
    \upvspacefig
    \includegraphics[width=\linewidth]{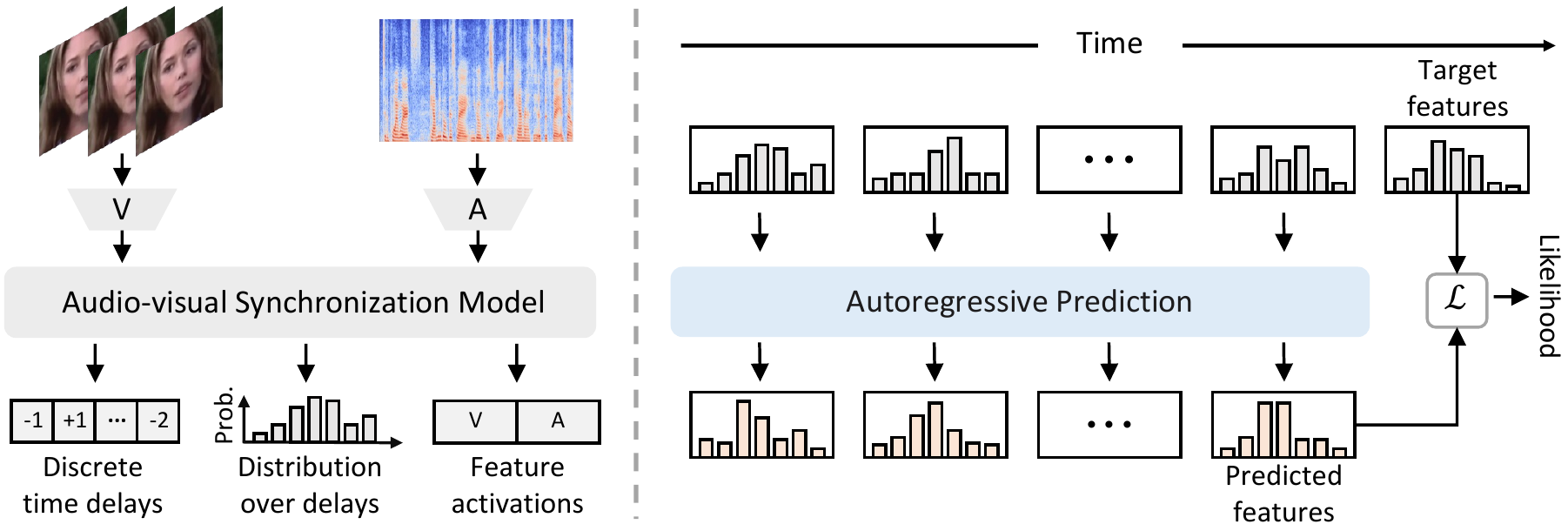}
    \begin{flushleft}
        \vspace{-3mm}
        \hspace{6mm} (a) Synchronization feature extraction \hspace{45mm} (b) Anomaly detection
         \vspace{-3mm}
    \end{flushleft}    
    \caption{\textbf{Audio-visual anomaly detection model.}
    (a) We extract a feature set from an audio-visual synchronization network: the number of frames of delay between video frames and sound, the distribution over delays at each frame, and feature activations from the audio and visual subnetworks. (b) We train an autoregressive Transformer model to assign probabilities to synchronization features. At test time, we flag low probability examples.}
    \label{model_overview}
\end{figure*}
\label{sec:method}

We formulate the problem of detecting manipulated videos as an anomaly detection problem. We model the distribution of audio-visual examples, then flag examples that have low probability. If we were to fit a model on the raw data, then this would be a very challenging learning task. Instead, we learn the distribution over a feature set that conveys subtle properties that are unlikely to be accurately captured in manipulated video. %

\subsection{Estimating audio-visual synchronization}\label{sync_estimate}

We obtain our feautres from a network that performs audio-visual synchronization~\cite{chung2019perfect,chung2016out,owens2018audio,chen2021audio}. We use the model of Chen \etal~\cite{chen2021audio}. We learn a function $\phi(V_i, A_j)$ that indicates how likely video clip $V_i$ temporally co-occurs with audio clip $A_j$. We estimate the synchronization score $S(i,j)$ of all audio-visual pairs in a temporal window: %
\begin{equation}
    S(i,j) = \frac{\exp{\left(\phi(V_i, A_j)\right)}}
    {\sum_{k=i-\tau}^{i+\tau}\exp{\left(\phi(V_i, A_k)\right)}},
    \label{sync_score}
\end{equation}
where $\tau$ is maximum time difference between two streams, and $\phi(V_i, A_j) = h\left(g_v(V_i), g_a(A_j)\right)$ is calculated using late fusion by a visual encoder $g_v$, audio encoder $g_a$ and the fusion module $h$. We also interpret $S(i,j)$ as synchronization probability. We maximize the synchronization of true audio-visual pairs $(V_{i},A_{i})$ using the InfoNCE loss~\cite{oord2018representation}: 
\begin{equation}
    \mathcal{L}_{sync} = - \frac{1}{T} \sum_{i=1}^{T}  \log S(i, i),
    \label{infonce}
\end{equation}
for a video of length $T$. We provide details about the architecture and training procedure in \supparxiv{the supplement}{\cref{appendix:imple}}.

After training, we can use the learned model to obtain a feature set for anomaly detection. For example, we can use the rows of $S$, which provide a probability distribution over possible alignments between video clips and audio clips.

\subsection{Audio-visual anomaly detection}\label{av_anomaly}

We use our learned model to obtain a feature set for anomaly detection.  We learn the distribution of these features on a training set of real videos. Then at test time, videos with low probability will be flagged as potential fakes. We now explore two key design decisions that go into such a system: what feature set to use, and how the distribution is learned. 

Given features for each frame, we learn a distribution $p_\theta(\bx_1, \bx_2, \dots, \bx_N)$. We generally use autoregressive models to learn this distribution, given their success in modeling complex distributions~\cite{brown2020language,yu2022scaling}. These models take the form: 
\begin{equation}\label{chain_decom}
    p_\theta(\mathbf{x}_{1},\mathbf{x}_{2},\cdots,\mathbf{x}_{N}) = \prod_{i=0}^{N-1}p_\theta(\mathbf{x}_{i+1}|\mathbf{x}_{1},\cdots, \mathbf{x}_{i}).
\end{equation}
We train a model $\hat \bx_{i+1} = f_\theta(\bx_1, \bx_2, \dots, \bx_i)$ that estimates the features of the next frame, given all of the features from the previous frames. Maximizing the log probability can be posed as minimizing a per-frame loss, $L$:
\begin{equation}
\mathcal{L} = \sum_{i=1}^{N} L(\hat \bx_{i}, \bx_i).
\label{eq:loss}
\end{equation}

 We now describe different formulations of the loss function $L$, the feature representation $\bx_i$. In each case, we implement $f_\theta$ as a Transformer~\cite{vaswani2017attention}.

\mypar{Discrete time delays.} 
We first consider a simple model that uses discrete {\em time delay} as our features, following the success of autoregressive models for fitting discrete data~\cite{van2017neural, razavi2019generating, esser2021taming}. Inspired by work on time delay estimation~\cite{knapp1976generalized,chen2022sound}, for every video frame, we estimate how far ahead (or behind) it appears to be from the audio signal. For each frame, we set $\mathbf{x}_{i}$ to be the time delay with the highest probability, \ie, $x_{i} = \argmax_j(S(i, j))$. We then set $L$ to be the cross entropy loss between the ground truth and predicted time delay. This amounts to solving a categorization problem with $2\tau + 1$ possible labels for each frame.

\mypar{Distributions over delays.}\label{kl-model} 
While discrete time delays are straightforward to represent in the model, they discard important information, such as when there is ambiguity in the delay.
We, therefore, propose a model that directly predicts the entries of the time delay distribution. 
We set the features $\bx_i$ to be the rows of $S$, i.e. the probability of each possible delay, and use cross entropy loss:
\begin{equation}
    L(\hat \bx_i, \bx_i) = - \sum_j^{2\tau+1} x_{i,j} \log(\hat x_{i,j}).
\end{equation}

We constrain the predictions made by our model $f_\theta$ to sum to 1 by applying a softmax.

\mypar{Audio-visual network activations.}
The feature activations within the audio-visual synchronization network convey information about the time delay. We, therefore, ask whether these activations can be directly used as features for anomaly detection. We concatenate the representations of the visual and audio subnetworks, $g_v$ and $g_a$. To provide a straightforward comparison with the time delay distribution model, we reduce the dimensionality of the features by projecting them onto the top $2\tau + 1$ principal components, following other work in autoregressive models of features~\cite{ramesh2022hierarchical}. We use squared distance as our loss: $L\left(\hat{\mathbf{x}}_i, \mathbf{x}_i\right) = \lVert \bx_i - \hat \bx_i \rVert^2$.

\section{Results}
We evaluate the different variations of our model on a variety of video forensics tasks. 

\subsection{Implementation details}
\label{implementation}
\paragraph{Synchronization model.} 

Following Chen \etal~\cite{chen2021audio}, we use ResNet-18 2D+3D\cite{he2016deep, hara2018can} as the visual encoder, using 5 frames~(25 fps) as input. The audio encoder uses VGG-M~\cite{chatfield2014return} and extracts features from 0.2s audio clips~(16kHz).
We fuse audio and visual data using a Transformer that has 3 standard Transformer encoder blocks~\cite{vaswani2017attention}, 4 attention heads, and 512 channels. We train using the cropped faces provided by each dataset. Please see \supparxiv{the supplement}{\cref{appendix:imple}} for details.

\mypar{Anomaly detection model.} 
We use a decoder-only autoregressive Transformer~\cite{liu2018generating,esser2021taming,radford2019language} to learn the distribution over synchronization features. We use 2 decoder blocks~\cite{vaswani2017attention}, each with 16 attention heads and 256 channels. 
For models that use time delay or continuous distribution, we set the maximum delay to be $\tau=15$ frames, resulting in the distribution $S_i \in \mathbb{R}^{31}$ for each video frame. We use sequences of length $N=50$ from 2.0s video.

\mypar{Hyperparameters.} 
We resample videos to 25 fps and audios to 16kHz. We represent audio segments as mel spectrograms of size $21\times80$ by short-time Fourier transform~(STFT) with 80 mel filter banks, a hop length of 160, and a window size of 320. Please see more details in \supparxiv{the supp}{\cref{appendix:imple}}.

\begin{table*}[t!]
\renewcommand{\arraystretch}{1.1}
\centering
\upvspacefig
\resizebox{\textwidth}{!}{
\begin{tabular}{clclccccccccccc:cc}
\toprule

&  &   & \multirow{4}{*}{\begin{tabular}[c]{c}Pretrained\\ dataset\end{tabular}} & \multicolumn{13}{c}{Category}
\\
\cmidrule(lr){5-17} 
    &        Method       &        Modality                   &                                & \multicolumn{2}{c}{RVFA} & & \multicolumn{2}{c}{FVRA-WL} & \multicolumn{2}{c}{FVFA-FS} & \multicolumn{2}{c}{FVFA-GAN} & \multicolumn{2}{c}{FVFA-WL}& \multicolumn{2}{c}{AVG-FV}
    \\
\cmidrule(lr){5-6} \cmidrule(lr){8-9} \cmidrule(lr){10-11} \cmidrule(lr){12-13} \cmidrule(lr){14-15} \cmidrule(lr){16-17}
&                                            &                             &                                                                               & AP         & AUC   &       & AP            & AUC            & AP            & AUC            & AP             & AUC            & AP            & AUC         & AP            & AUC       \\
\midrule 
\parbox[t]{4mm}{\multirow{5}{*}{\rotatebox[origin=c]{90}{Supervised}}} 
 & Xception~\cite{rossler2019faceforensics++}                         &     $\mathcal{V}$                    &            ImageNet~\cite{deng2009imagenet}                                                           &   --         &     --     &    &   88.2&  88.3     &  92.3     &     93.5     & 67.6       &    68.5    &91.0     & 91.0  &   84.8  &  85.3       \\
& LipForensics~\cite{haliassos2021lips}                         &     $\mathcal{V}$                    &            LRW~\cite{Chung16}                                                               &   --         &     --  &    &    \textbf{97.8}       &       \textbf{97.7}         &       99.9      &      99.9         &        61.5      &     68.1      &        98.6   &     98.7  & 89.4 & 91.1  \\
& AD DFD~\cite{zhou2021joint}              &     $\mathcal{A} \mathcal{V}$                                           &     Kinetics~\cite{kay2017kinetics}                                                                          &       \textbf{74.9}       &       \textbf{73.3} &   &    97.0         &     97.4          &      99.6       &     99.7         &         58.4   &   55.4       &        \textbf{100.}    &      \textbf{100.}    & 88.8 & 88.1     \\
& FTCN~\cite{zheng2021exploring}                                   &        $\mathcal{V}$                      &          --                                                                     &    --        &   --       &   &   96.2    &      97.4       &    \textbf{100.}          &      \textbf{100.}        &      77.4       &       78.3    &    95.6       &    96.5    & 92.3 &   93.1    \\
& RealForensics~\cite{haliassos2022leveraging}                         &         $\mathcal{V}$                     & LRW~\cite{Chung16}                                                                          &   --        &  --   &  &   88.8      &     93.0          &      99.3       &         99.1    &           \textbf{99.8}     &       \textbf{99.8}      &   93.4          &     96.7  & \textbf{95.3} & \textbf{97.1}       \\
\cmidrule(lr){1-17}
\parbox[t]{4mm}{\multirow{4}{*}{\rotatebox[origin=c]{90}{Unsupervised}}} 
& AVBYOL~\cite{haliassos2022leveraging,grill2020bootstrap}               &     $\mathcal{A} \mathcal{V}$                                                 & LRW~\cite{Chung16}                                                                         &    50.0        &      50.0 &  & 73.4  &       61.3      &     88.7     &  80.8 &  60.2     &      33.8            &   73.2         & 61.0 & 73.9  & 59.2 \\
& VQ-GAN ~\cite{esser2021taming}                                 &      $\mathcal{V}$                   & LRS2~\cite{Afouras18c}                                                                        & --       &     --  &   &  50.3     &   49.3       & 57.5  &   53.0      &    49.6      &     48.0    &     62.4        &  56.9  &  55.0 & 51.8  \\
\cdashlinelr{2-17}
& Ours                &                 $\mathcal{A} \mathcal{V}$                                  & LRS2~\cite{Afouras18c}                                                                         &      62.4    &  71.6    &    &     \textbf{93.6}         &     \textbf{93.7}    &  \textbf{95.3} &    \textbf{95.8}     &       \textbf{94.1}       &       \textbf{94.3}       &   \textbf{93.8}           & \textbf{94.1}  &\textbf{94.2} & \textbf{94.5}  \\
& Ours                &                 $\mathcal{A} \mathcal{V}$                                  & LRS3~\cite{afouras2018lrs3}        &      \textbf{70.7}      &     \textbf{80.5} &     &     91.1         &    93.0         &   91.0   &  92.3        &   91.6           &      92.7       &   91.4           & 93.1   & 91.3 & 92.8\\ 
\bottomrule
\end{tabular}
}
\caption{\textbf{Manipulation detection on FakeAVCeleb.} We report AP scores $(\%)$ and AUC scores $(\%)$, following the evaluation protocol of Haliassos et al.~\cite{haliassos2021lips}, in which supervised methods are evaluated on unseen manipulation types (unsupervised methods are not trained with labels and fake examples). We report results with combinations of real/fake video/audio, using different manipulation algorithms. We report the average performance over four fake video~(FV) categories in AVG-FV. We retrained all supervised models on FakeAVCeleb~\cite{NEURIPS_DATASETS_AND_BENCHMARKS2021_d9d4f495}.}
\label{comparison_with_others}
\end{table*}

\subsection{Dataset}
We train our model on real, unlabeled speech video, and evaluate it on forensics datasets.

\mypar{Training datasets.} We train our models on Lip Reading Sentences 2~(LRS2, 97k videos)~\cite{Afouras18c} and Lip Reading Sentences 3~(LRS3, 120k videos)~\cite{afouras2018lrs3}. The videos in each contain tightly cropped face tracks. We divide each dataset into 3 splits and train the audio-visual synchronization model and the autoregressive model on different splits.

\mypar{Evaluation datasets.} 
We evaluate on two video forensics datasets, spanning several different types of manipulations that change the speech and face of a human speaker. 
\textbf{FakeAVCeleb}~\cite{NEURIPS_DATASETS_AND_BENCHMARKS2021_d9d4f495}, which is derived from VoxCeleb2~\cite{chung2018voxceleb2}. This dataset contains 500 real videos and 19,500 fake videos manipulated by Faceswap~\cite{korshunova2017fast}, FSGAN~\cite{nirkin2019fsgan}, and Wav2Lip~\cite{prajwal2020lip}, and fake sounds that are generated by SV2TTS~\cite{jia2018transfer}. The examples in the dataset contain different combinations of these manipulations. We use the dataset's provided face crops. We sample 2400 videos (400 real videos and 2000 fake videos) as train/val splits and 600 videos (100 real videos and 500 fake videos) as test split. We note that our method does not use any videos from train/val splits, since it is trained from another dataset (LRS2 or LRS3). %
Second, we evaluate on \textbf{KoDF}~\cite{kwon2021kodf}, a large-scale Korean-language deepfake detection dataset. It contains 62,166 real videos and 175,776 fake videos, where fake videos are generated by 6 synthesized methods: FaceSwap~\cite{faceswap}, FSGAN~\cite{nirkin2019fsgan}, DeepFaceLab~\cite{perov2020deepfacelab}, FOMM~\cite{siarohin2019first}, ATFHP~\cite{yi2020audio} and Wav2Lip~\cite{prajwal2020lip}. We extract faces by using face detection~\cite{zhang2017s3fd} and alignment~\cite{bulat2017far}.

\subsection{Evaluation methods}
Following common practice~\cite{wang2019cnn, rossler2019faceforensics++, li2020face, chai2020makes, qian2020thinking, haliassos2021lips, zheng2021exploring, haliassos2022leveraging, kwon2021kodf, NEURIPS_DATASETS_AND_BENCHMARKS2021_d9d4f495}, we evaluate using average precision~(AP) and AUC. These evaluation metrics are widely used for cross-dataset generalization and unsupervised models since they avoid the need to threshold the predictions. We compare our approach to both supervised and self-supervised methods at the video level. Unless otherwise stated, we use time delay distributions as our feature set (\cref{kl-model}). 

\mypar{Supervised methods.}
For supervised methods, we retrain several state-of-the-art detectors on FakeAVCeleb~\cite{NEURIPS_DATASETS_AND_BENCHMARKS2021_d9d4f495}: 1)~{\bf Xception}~\cite{rossler2019faceforensics++}: a popular baseline for forensics detection; 2)~{\bf LipForensics}~\cite{haliassos2021lips}: a detector is built on high-level semantic embeddings of mouth and targets irregularities in mouth movements; 3)~{\bf AD DFD}~\cite{zhou2021joint}: a multimodal detector with audio and video branches, utilizes audio-visual synchronization signal implicitly for detection; 4)~{\bf FTCN}~\cite{zheng2021exploring}: a video forensics detector leverages temporal incoherence to boost generalization capability; 5)~{\bf RealForensics}~\cite{haliassos2022leveraging}: it first pretrains the network by audio-visual BYOL~\cite{grill2020bootstrap} framework and then finetunes the pretrained model on forensics datasets by multi-task learning to obtain robust and general face forgery detection.

\mypar{Self-supervised methods.}
Since we are not aware of any existing methods that consider self-supervised speech video forensics, we adapt two existing methods to the task. First, we consider an audio-visual contrastive learning model, which we call  {\bf AVBYOL}, that learns to determine whether the visual and audio streams of a video do (or do not) match, an approach that has been used as a part of other audio-visual forensics models~\cite{haliassos2022leveraging,cozzolino2022audio}. We adapt the model of Haliassos \etal~\cite{haliassos2022leveraging}, which uses BYOL~\cite{grill2020bootstrap} to learn a joint audio-visual embedding for pretraining. Instead of pretraining, we directly use the model's audio-visual similarity score to flag fake examples.
Second, we use a generative model {\bf VQGAN}~\cite{esser2021taming}, trained on LRS2~\cite{Afouras18c}, for anomaly detection. VQGAN converts an image into a sequence of discrete codes, then uses an autoregressive Transformer to learn the distribution of codes. We use the code's log likelihood, averaged over each video frame, for anomaly detection.

\subsection{Evaluation}
In real-world scenarios, the deployed detectors are expected to recognize fake videos manipulated by unseen techniques. Thus, following the standard procedure used in \cite{haliassos2022leveraging, haliassos2021lips, zheng2021exploring, zhou2021joint}, we conduct the experiment to evaluate the cross-manipulation generalization ability of our model on the FakeAVCeleb dataset~\cite{NEURIPS_DATASETS_AND_BENCHMARKS2021_d9d4f495} which videos are manipulated in various ways. Since our approach and other self-supervised baselines learn from real, unmanipulated videos and perform zero-shot fake video detection, all the fake videos during evaluation are considered as manipulated by unseen methods.

We split FakeAVCeleb dataset~\cite{NEURIPS_DATASETS_AND_BENCHMARKS2021_d9d4f495} into five categories based on the manipulation methods and manipulated modalities: 
1)~{\bf RVFA}: real video with fake audio by SV2TTS~\cite{jia2018transfer}; 
2)~{\bf FVRA-WL}: real audio with fake video by Wav2Lip~\cite{prajwal2020lip}; 
3)~{\bf FVFA-WL}: fake video by Wav2Lip~\cite{prajwal2020lip}, and fake audio by SV2TTS~\cite{jia2018transfer};
4)~{\bf FVFA-FS}: fake video by Faceswap~\cite{korshunova2017fast} and Wav2Lip\cite{prajwal2020lip}, and fake audio by SV2TTS~\cite{jia2018transfer};
5)~{\bf FVFA-GAN}: fake video by FaceswapGAN~\cite{nirkin2019fsgan} and Wav2Lip\cite{prajwal2020lip}, and fake audio by SV2TTS~\cite{jia2018transfer}.
For supervised methods, we hold out the evaluated category and train the models on the remaining categories. Note that some approaches are only able to detect the manipulation on a certain modality, we do not report their performance on the categories with the manipulation only on the other modalities (since they can not differentiate real and fake videos).

We show our results in \cref{comparison_with_others}. Our method substantially outperforms both self-supervised methods AVBYOL~\cite{haliassos2022leveraging, grill2020bootstrap} and VQGAN~\cite{esser2021taming} on each category by a large margin. More importantly, our method works on par with or outperforms some supervised methods on certain categories, especially FVFA-GAN, even though our method does not use any labeled supervision or fake examples.  Moreover, our method has quite consistent performances and it can achieve AP over 90\% in the most of categories. While Xception~\cite{rossler2019faceforensics++}, LipForensics~\cite{haliassos2021lips}, AD DFD~\cite{zhou2021joint} and  FTCN~\cite{zheng2021exploring} work well on 75\% of the settings, there are settings where performance collapses to near-chance~(\eg, AD DFD~\cite{zhou2021joint} on FVFA-GAN). Interestingly, two self-supervised baselines struggle to detect fake videos, perhaps because both models do not necessarily capture the subtle information that would be needed to detect manipulations. In addition, 
VQGAN~\cite{esser2021taming} compresses the visual signal using a codebook, which might drop the artifact clues and harm the detection performance. Moreover, our model trained on LRS2~\cite{Afouras18c} works on par with the one trained on LRS3~\cite{afouras2018lrs3}, indicating that our method's performance is not tied to a single training set.

\begin{table}[t!]
\centering
\upvspacefig
\resizebox{1\columnwidth}{!}{
\begin{tabular}{llccc}
\toprule
\multicolumn{2}{c}{\multirow{2}{*}{Method}} & \multirow{2}{*}{Modality} &\multicolumn{2}{c}{KoDF~\cite{kwon2021kodf}} \\
\cmidrule(lr){4-5}
\multicolumn{2}{c}{}           &             & AP          & AUC         \\
\midrule
\multirow{5}{*}{\shortstack[c]{Supervised \\ (transfer)}}
 & Xception~\cite{rossler2019faceforensics++}   %
   & $\mathcal{V}$  &   76.9  &  77.7\\
                            & LipForensics~\cite{haliassos2021lips}   &  $\mathcal{V}$ & 89.5       &    86.6       \\
                            & AD DFD~\cite{zhou2021joint}   & $\mathcal{AV}$      &   79.6         &   82.1       \\
                            & FTCN~\cite{zheng2021exploring} & $\mathcal{V}$           &     66.8       &     68.1       \\
                            & RealForensics~\cite{haliassos2022leveraging}   &        $\mathcal{V}$ &  \textbf{95.7}    &          \textbf{93.6}   \\
\midrule
\multirow{3}{*}{Unsupervised}                  
& AVBYOL~\cite{haliassos2022leveraging, grill2020bootstrap} &$\mathcal{AV}$   & 74.9     & 78.9 \\

& VQ-GAN ~\cite{esser2021taming} & $\mathcal{V}$   &   46.8      &   45.5     \\

\cdashlinelr{2-5}
& Ours       & $\mathcal{AV}$    &        \textbf{87.6}    &      \textbf{86.9 }        \\

\bottomrule
\end{tabular}
}
\caption{\textbf{Generalization to Korean speech.} AP scores ($\%$) and AUC scores ($\%$) are reported on KoDF dataset~\cite{kwon2021kodf}. Supervised methods are trained on FakeAVCeleb dataset~\cite{NEURIPS_DATASETS_AND_BENCHMARKS2021_d9d4f495}. Ours is trained on LRS2~\cite{Afouras18c}. Best results are in \textbf{bold}.}
\label{kodf_result}
\end{table}

\mypar{Cross-dataset generalization.} 
We also evaluate the generalization capability of our model by evaluating it on the KoDF dataset~\cite{kwon2021kodf}, following \cite{haliassos2022leveraging, haliassos2021lips, zheng2021exploring, zhou2021joint}. We focus on the audio-driven synthesis examples in the dataset, where videos are manipulated by ATFHP~\cite{yi2020audio} or Wav2Lip~\cite{prajwal2020lip}, and randomly selected 100 real and 100 fake videos. We train the supervised models on FakeAVCeleb~\cite{NEURIPS_DATASETS_AND_BENCHMARKS2021_d9d4f495} to evaluate their generalization ability. Many of these training videos share the same technique used in KoDF for synthesis~\cite{prajwal2020lip}. As the results are shown in \cref{kodf_result}, our approach obtains a comparable performance to many supervised methods. Although our system is trained on the English speech datasets, it still generalizes to KoDF~\cite{kwon2021kodf} dataset of Korean speech, perhaps because it learns low-level lip motion cues that are broadly useful. We provide more results in \supparxiv{the supplement}{\cref{appendix:crossdataset}}.

\begin{figure}[t!]
    \centering
    \upvspacefig
    \includegraphics[width=\linewidth]{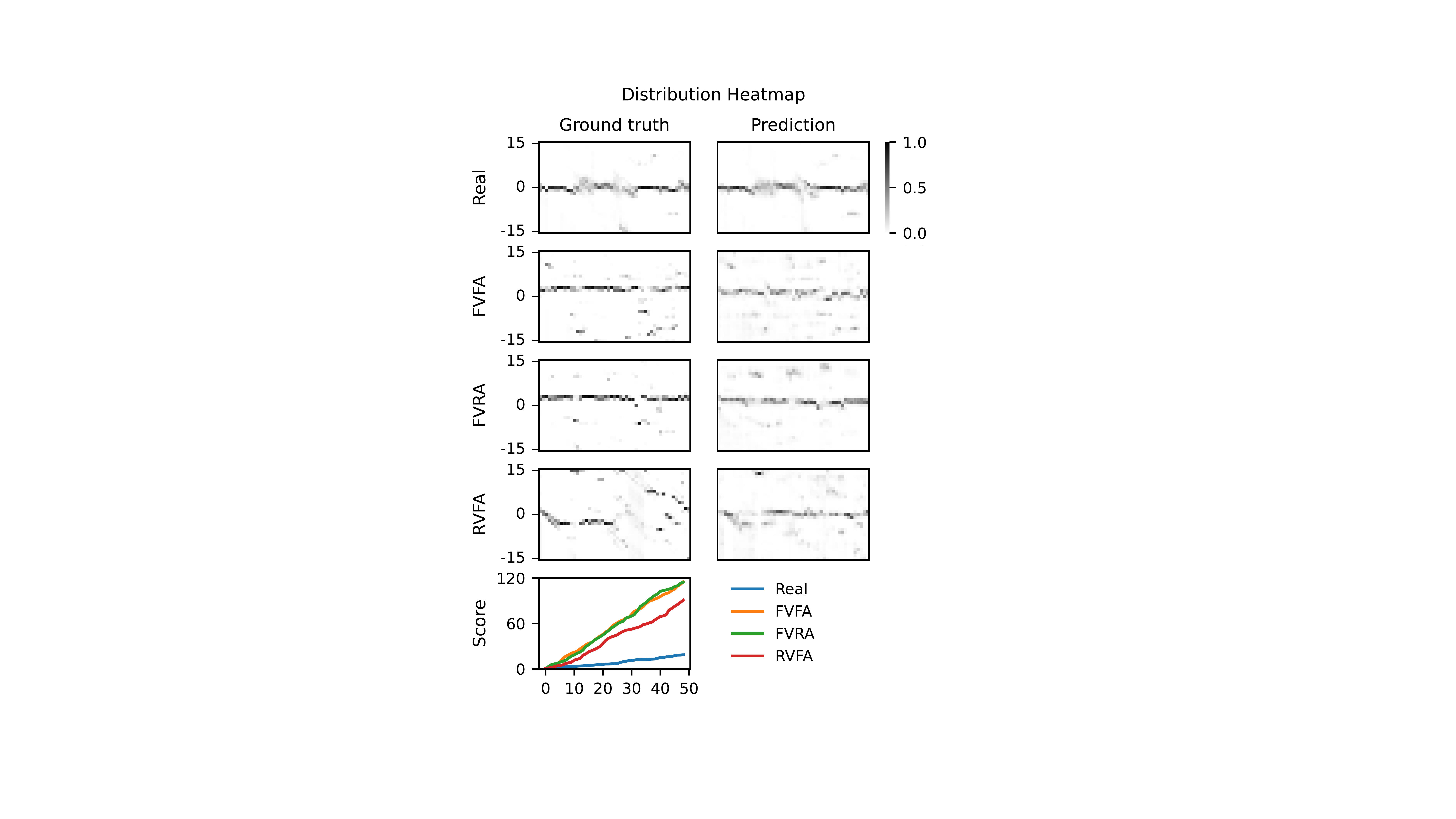}
    \caption{\textbf{Time delay distribution predictions for real vs. fake examples.} We visualize the time delay distributions from the synchronization model and predicted results generated by the autoregressive model for four random samples from different categories. Synchronization probabilities are in a range from 0~\includegraphics[width=0.10\linewidth, height=0.03\linewidth]{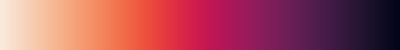}~1. We show the predictions of the autoregressive model when feeding it ground truth observations of the previous timesteps. We show cumulative prediction error (indicating the probability of being fake) for each sample over time steps in the last row. } %
    \label{heatmap}
\end{figure}

\begin{figure*}[t!]
    \centering
    \upvspacefig
 \includegraphics[width=0.95\linewidth]{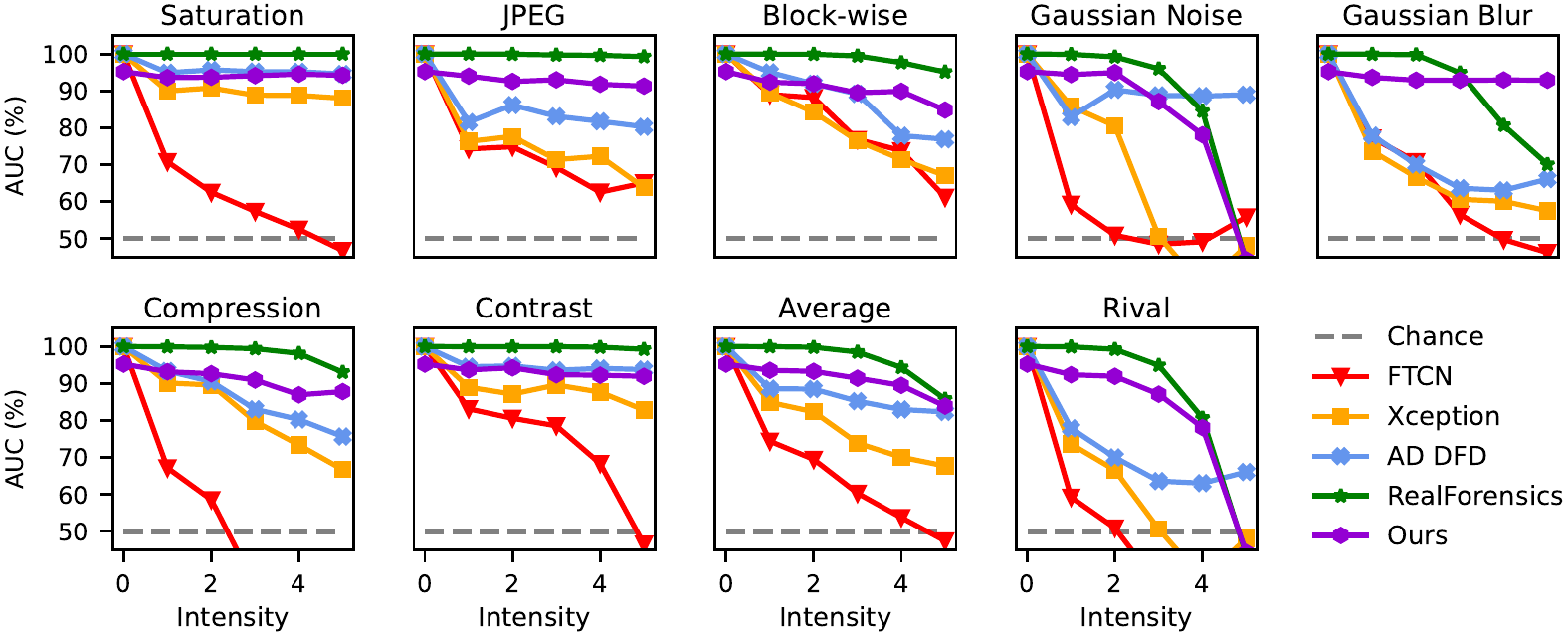}
    \caption{\textbf{Robustness to unseen perturbations.} AUC scores $(\%)$ of different detectors as a function of perturbation intensities. There are 6 intensity levels in total from~\cite{jiang2020deeperforensics}. ``Average'' represents the average over 7 perturbations under each intensity. ``Rival'' means we pick the worst performance across 7 perturbations under each intensity.}
    \label{robustness}
\end{figure*}

\mypar{Qualitative results.}
We visualize the ground truth and predicted time delay distributions generated by our autoregressive continuous time delay model (\cref{av_anomaly}) in \cref{heatmap}. We use the four main categories from FakeAVCeleb dataset~\cite{NEURIPS_DATASETS_AND_BENCHMARKS2021_d9d4f495}. For each one, we display a heat map indicating the predicted time delay distribution, using a model that obtains the ground truth distributions of the previous frames as input.
We also plot the cumulative prediction loss (Eq.~\ref{eq:loss}) over time. From \cref{heatmap}, we can see that our autoregressive model accurately predicts the ground truth for real video, which results in a lower score. For fake videos, we can find clear differences between ground truth and predicted time delay distribution, leading to higher prediction loss.

\subsection{Robustness to unseen perturbations}\label{subsec_robustness}

When the fake video is redistributed, it may undergo many types of postprocessing that result in corruption, making detection more difficult. Thus, it is important for forensics models to be robust to the types of postprocessing operations they may encounter in the wild. Following \cite{haliassos2021lips,zheng2021exploring,haliassos2022leveraging}, we use the set of visual perturbations proposed in \cite{jiang2020deeperforensics}: 1)~Color saturation change; 2)~Block-wise distortion; 3)~Color contrast change; 4)~Gaussian blur; 5)~Gaussian noise; 6)~JPEG compression; 7)~Video compression rate change. We set the intensity levels from 0 to 5 for each perturbation. 

We compare our model with four supervised methods XceptionNet~\cite{rossler2019faceforensics++}  FTCN~\cite{zhou2021joint}, AD DFD~\cite{zhou2021joint} and RealForensics~\cite{haliassos2022leveraging}. 
As shown in \cref{robustness}, our self-supervised model is overall more robust to unseen visual perturbations on average compared with these supervised methods, with the exception of RealForensics~\cite{haliassos2022leveraging}. This is also true when we consider ``worst case'' performance, by taking the minimum performance over all types of augmentation of a given intensity level. Interestingly, we obtain this performance even though our model is trained in a very different way from other works, suggesting that the feature set continues to convey useful information to the anomaly detection model, even in the presence of significant corruption.

\subsection{Feature set analysis}
\label{feature_analysis}
We evaluate the effectiveness of different feature sets used by our anomaly detection model. As described in \cref{av_anomaly}, we start with {\bf discrete time delays} as our feature representation and optimize the model with the cross entropy loss. Then, we use {\bf continuous time delay distributions} as representations instead, where we optimize models with different objective functions: 1) Soft CE: we use the time delay distribution as the target (akin to a ``soft'' label) and use the cross entropy loss (\cref{av_anomaly}); 2) CE: we map each distribution into one-hot encoding as the target by using $\argmax$ and employ the cross entropy loss; 3) BCE: we use the distribution as the target while treating each synchronization score $S(i,j)$ (\cref{sync_estimate}) within the same time step independently. We use the sigmoid function and binary cross entropy loss to train the model. We also use our network's {\bf feature activations} as in \cref{av_anomaly}: 1) audio-visual feature activations (activation-AV); 2) visual-only feature activations (activation-V). Besides, we consider using a combination of different feature sets where we concatenate {\bf continuous time delay distributions} and {\bf audio-visual feature activations}~(Act.-AV$+$dist.) as a new feature. Similar to audio-visual feature activations as in \cref{av_anomaly}, we use squared distance as the loss for the concatenation of these two types of feature sets. 

\begin{table}[t]
\centering
\resizebox{\columnwidth}{!}{
\begin{tabular}{lllcccc}
\toprule
\multirow{2}{*}{Model} &\multirow{2}{*}{Feature set} & \multirow{2}{*}{$\mathcal{L}$} & \multicolumn{2}{c}{AVG-ALL}&\multicolumn{2}{c}{AVG-FV} \\
\cmidrule(lr){4-5} \cmidrule(lr){6-7}
     &      &  & AP & AUC & AP& AUC\\
\midrule
Bayes &-  & -& 73.1 & 85.1 & 72.4 &  86.0 \\
\cdashlinelr{1-7}
\multirow{7}{*}{Ours} & discrete delay & CE &    80.8 &    86.5 & 80.0 & 86.6 \\
 & distribution &   CE &  84.8 &   87.9 & 90.3 & 92.2 \\
  &  distribution & BCE &  78.6 &   83.4   &80.5 &84.8 \\
   &  distribution & Soft CE &   \textbf{87.8} &  \textbf{90.0}  & \textbf{94.2}  & \textbf{94.5} \\
      &  activation-AV &  MSE &    86.5 & 87.1 &91.5  & 91.9 \\
        & Act.-AV$+$dist.& MSE &  85.5 &  87.0 &     90.0  & 91.3 \\
       &  activation-V &  MSE &    -- &  -- & 77.6  & 85.9 \\
    & discrete prob. &  -- &  83.4 &     86.9  & 88.6 &  91.1 \\
\bottomrule
\end{tabular}}
\caption{\textbf{Feature set analysis.} AP~($\%$) and AUC~($\%$) are reported on FakeAVCeleb~\cite{NEURIPS_DATASETS_AND_BENCHMARKS2021_d9d4f495} when using different feature sets. Best results are in \textbf{bold}. AVG-ALL means the average over all categories. AVG-FV represents the average over four fake video categories.}
\label{naive_bayes_comparison_main}
\end{table}

 We also compare with a simple model based on {\bf Naive Bayes} and discrete time delays. This model assumes that each frame's time delay is independent, and obtains a probability for the entire sequence by multiplying the probability of each frame's time delay. This amounts to simply detecting large misalignments since in practice the Naive Bayes model will assign probability solely based on the magnitude of each delay.

Finally, we consider a version of the model that autoregressively predicts the entire distribution of time delays, inspired by autoregressive models, such as PixelCNN~\cite{van2016conditional} that generate images in a raster scan order. We autoregressively predict each element of the 2D matrix $\hat{S}(i,j)$, where $\hat{S}(i,j)$ is created by vector quantizing the entries  of the synchronization probability $S(i,j)$ using $k$-means (see \supparxiv{supp.}{\cref{appendix:feature_set}} for details). 

\mypar{Analysis.} We evaluate each variant on FakeAVCeleb~\cite{NEURIPS_DATASETS_AND_BENCHMARKS2021_d9d4f495} and report results in \cref{naive_bayes_comparison_main}. These results suggest that all formulations achieve performance significantly better than chance, indicating that these feature sets are useful for anomaly detection. As in \cref{naive_bayes_comparison_main}, the time delay distribution model outperforms the discrete time delay model, suggesting that there is important information conveyed in the probability of unlikely delays. The autoregressive model that uses distribution as input and soft labels (soft CE) performs best since it forces the output prediction to match the distribution from the synchronization model. Interestingly, the model that uses audio-visual feature activations obtains performance close to that of the soft CE model, indicating that the networks' audio-visual features convey useful information.  Finally, the multimodal activation-AV model significantly outperforms the visual-only activation-V model, suggesting that having access to both modalities is useful for our anomaly detection model.

\subsection{Ablation study}

\begin{table}[t!]
\centering
\upvspacefig
\resizebox{0.89\columnwidth}{!}{
\begin{tabular}{llcc}
\toprule
\multirow{2}{*}{\begin{tabular}[c]{@{}l@{}}Synchronization\\ dataset\end{tabular}} &\multirow{2}{*}{\begin{tabular}[c]{@{}l@{}}Auto-regression\\ dataset\end{tabular}}& \multicolumn{2}{c}{AVG}\\
\cmidrule(lr){3-4}
     &   & AP           & AUC          \\
\midrule
\multirow{3}{*}{LRS2~\cite{Afouras18c}}  & LRS2~\cite{Afouras18c} & \textbf{87.8}          &      90.0       \\
  & LRS3~\cite{afouras2018lrs3} &   85.0       &   89.6 \\       
 & LRS2~\cite{Afouras18c}$+$LRS3~\cite{afouras2018lrs3} &  85.1       &   89.9 \\     
\cdashlinelr{1-4}
\multirow{3}{*}{LRS3~\cite{afouras2018lrs3}} & LRS2~\cite{Afouras18c}   &   86.6        & 89.0 \\

& LRS3~\cite{afouras2018lrs3}   &    87.2       &  90.3 \\
& LRS2~\cite{Afouras18c}$+$LRS3~\cite{afouras2018lrs3} &  87.2      &   {\bf 90.6} \\    

\bottomrule
\end{tabular}
}
\caption{\textbf{Dataset ablation.} AP scores ($\%$) and AUC scores ($\%$) are reported on FakeAVCeleb~\cite{NEURIPS_DATASETS_AND_BENCHMARKS2021_d9d4f495} dataset by using different datasets to train synchronization model or atuoregressive model. Best results are in \textbf{bold}.}
\label{combined_dataset_main}
\end{table}

\paragraph{Different training dataset.}
\label{training_dataset}
We ask how the choice of dataset affects the quality of the model. To test this, we train our synchronization and autoregressive models on different datasets to analyze the generalization abilities of each component, \ie, training the synchronization model on LRS2/LRS3 and training the autoregressive model on LRS3/LRS2 or LRS2$+$LRS3 with the same hyperparameters.  As shown in \cref{combined_dataset_main}, there is no significant performance change when we train these two components on different combinations of datasets, including when they are trained on the same dataset. This suggests that the distribution of time delay predictions may be stable between these speech video datasets.

\begin{figure}[t!]
    \centering
    \upvspacefig
    \includegraphics[width=0.9\linewidth]{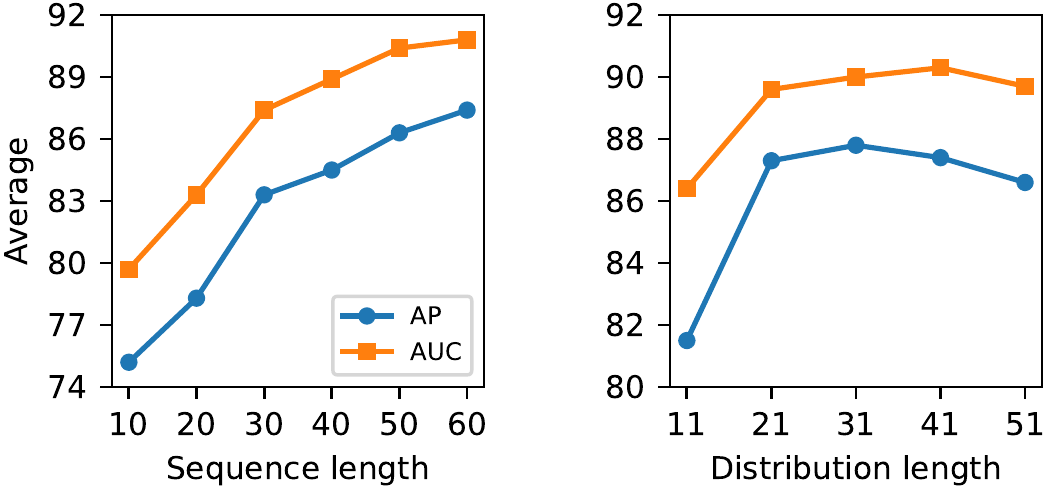}
    \caption{\textbf{Hyperparameter ablation.} 
    We evaluate with different input sequence lengths for our autoregressive model on FakeAVCeleb (left), and study the effect of the time delay distribution's maximum temporal offset (right).} 
    \label{clip_size_fig}
\end{figure}

\mypar{Influence of sequence length.}
To explore the influence of input sequence length for the autoregressive model, we sample the same amount of training videos for sequence lengths $N$ of 10, 20, 30, 40, 50, and 60, and keep other hyperparameters the same. We test these models on FakeAVCeleb dataset~\cite{NEURIPS_DATASETS_AND_BENCHMARKS2021_d9d4f495}. \cref{clip_size_fig} shows that as the sequence length increases, the performance increases with it. %

\mypar{Effect of time delay distribution maximum offset.}
We also study how the length of time delay distribution would affect the performance of the autoregressive model with distribution over delays. We experiment with maximum offset $\tau \in \left \{ 5, 10, 15, 20, 25 \right \}$ resulting in the delay distribution length of $\{11, 21, 31, 41, 51\}$. We test these models on the FakeAVCeleb dataset~\cite{NEURIPS_DATASETS_AND_BENCHMARKS2021_d9d4f495}. \cref{clip_size_fig} shows that as distribution length increases, the performance first increases, after which point results plateau or slightly decrease. This may be due to the fact that when considering larger ranges of offsets, the distribution spreads over a large number of unlikely possibilities, making important information less apparent after normalization.

\section{Conclusion}
We have proposed a method for detecting video manipulation by self-supervised anomaly detection. To do this, we create novel feature sets that convey audio-visual synchronization. We then show that fake videos can be detected by flagging examples with unlikely sequences of these features, according to a learned distribution. Our model obtains strong performance on the FakeAVCeleb and KoDF datasets, despite the fact that it was trained only on real video. It also obtains robustness to visual postprocessing operations and to videos containing other spoken languages. We see our work as opening in two directions. The first is in posing forensics as an anomaly detection problem with a self-supervised feature set. While we have proposed one such model, based on autoregressive sequence models, the field of anomaly detection offers many possible future approaches. The second direction is in developing new feature sets that are well-suited to forensics problems, beyond the synchronization features used in this work.

\mypar{Limitations and Broader Impacts. } Our work provides methods that can potentially be applied to detecting malicious video manipulations and disinformation. While we have shown that our model is capable of detecting several types of fake video, there may be other techniques that our model fails to detect. In particular, due to the design of our use of synchronization-based features, our model is not well suited to detecting manipulations that leave the synchronization between motion and sound relatively consistent, such as those that change a speaker's appearance without significantly changing the motion of their mouth. 

\mypar{Acknowledgements.} We thank David Fouhey, Richard Higgins, Sarah Jabbour, Yuexi Du, Mandela Patrick, Deva Ramanan, Haochen Wang, and Aayush Bansal for helpful discussions. This work was supported in part by DARPA Semafor. The views, opinions and/or findings expressed are those of the authors and should not be interpreted as representing the official views or policies of the Department of Defense or the U.S. Government.

{\small
\bibliographystyle{ieee_fullname}
\bibliography{avforensics}

\begin{thebibliography}{100}\itemsep=-1pt

\bibitem{faceswap}
Faceswap. https://github.com/deepfakes/faceswap, 2022.

\bibitem{afouras2021selfsupervised}
Triantafyllos Afouras, Yuki~M. Asano, Francois Fagan, Andrea Vedaldi, and
  Florian Metze.
\newblock Self-supervised object detection from audio-visual correspondence,
  2021.

\bibitem{Afouras18c}
T. Afouras, J.~S. Chung, A. Senior, O. Vinyals, and A. Zisserman.
\newblock Deep audio-visual speech recognition.
\newblock In {\em arXiv:1809.02108}, 2018.

\bibitem{afouras2018lrs3}
Triantafyllos Afouras, Joon~Son Chung, and Andrew Zisserman.
\newblock Lrs3-ted: a large-scale dataset for visual speech recognition.
\newblock {\em arXiv preprint arXiv:1809.00496}, 2018.

\bibitem{afouras2020asr}
Triantafyllos Afouras, Joon~Son Chung, and Andrew Zisserman.
\newblock Asr is all you need: Cross-modal distillation for lip reading.
\newblock In {\em ICASSP 2020-2020 IEEE International Conference on Acoustics,
  Speech and Signal Processing (ICASSP)}, pages 2143--2147. IEEE, 2020.

\bibitem{Afouras20b}
Triantafyllos Afouras, Andrew Owens, Joon~Son Chung, and Andrew Zisserman.
\newblock Self-supervised learning of audio-visual objects from video.
\newblock In {\em European Conference on Computer Vision}, 2020.

\bibitem{agarwal2020detecting}
Shruti Agarwal, Hany Farid, Ohad Fried, and Maneesh Agrawala.
\newblock Detecting deep-fake videos from phoneme-viseme mismatches.
\newblock In {\em Proceedings of the IEEE/CVF conference on computer vision and
  pattern recognition workshops}, pages 660--661, 2020.

\bibitem{alcazar2021maas}
Juan~Le{\'o}n Alc{\'a}zar, Fabian Caba, Ali~K Thabet, and Bernard Ghanem.
\newblock Maas: Multi-modal assignation for active speaker detection.
\newblock In {\em Proceedings of the IEEE/CVF International Conference on
  Computer Vision}, pages 265--274, 2021.

\bibitem{arandjelovic2018objects}
Relja Arandjelovic and Andrew Zisserman.
\newblock Objects that sound.
\newblock In {\em Proceedings of the European conference on computer vision
  (ECCV)}, pages 435--451, 2018.

\bibitem{asano2020labelling}
Yuki Asano, Mandela Patrick, Christian Rupprecht, and Andrea Vedaldi.
\newblock Labelling unlabelled videos from scratch with multi-modal
  self-supervision.
\newblock {\em Advances in Neural Information Processing Systems},
  33:4660--4671, 2020.

\bibitem{bengio2000neural}
Yoshua Bengio, R{\'e}jean Ducharme, and Pascal Vincent.
\newblock A neural probabilistic language model.
\newblock {\em Advances in neural information processing systems}, 13, 2000.

\bibitem{bianchi2012image}
Tiziano Bianchi and Alessandro Piva.
\newblock Image forgery localization via block-grained analysis of jpeg
  artifacts.
\newblock {\em IEEE Transactions on Information Forensics and Security},
  7(3):1003--1017, 2012.

\bibitem{bondi2017tampering}
Luca Bondi, Silvia Lameri, David Guera, Paolo Bestagini, Edward~J Delp, Stefano
  Tubaro, et~al.
\newblock Tampering detection and localization through clustering of
  camera-based cnn features.
\newblock In {\em CVPR Workshops}, volume~2, 2017.

\bibitem{brown2020language}
Tom Brown, Benjamin Mann, Nick Ryder, Melanie Subbiah, Jared~D Kaplan, Prafulla
  Dhariwal, Arvind Neelakantan, Pranav Shyam, Girish Sastry, Amanda Askell,
  et~al.
\newblock Language models are few-shot learners.
\newblock {\em Advances in neural information processing systems},
  33:1877--1901, 2020.

\bibitem{bulat2017far}
Adrian Bulat and Georgios Tzimiropoulos.
\newblock How far are we from solving the 2d \& 3d face alignment problem? (and
  a dataset of 230,000 3d facial landmarks).
\newblock In {\em International Conference on Computer Vision}, 2017.

\bibitem{chai2020makes}
Lucy Chai, David Bau, Ser-Nam Lim, and Phillip Isola.
\newblock What makes fake images detectable? understanding properties that
  generalize.
\newblock In {\em European conference on computer vision}, pages 103--120.
  Springer, 2020.

\bibitem{chatfield2014return}
Ken Chatfield, Karen Simonyan, Andrea Vedaldi, and Andrew Zisserman.
\newblock Return of the devil in the details: Delving deep into convolutional
  nets.
\newblock {\em arXiv preprint arXiv:1405.3531}, 2014.

\bibitem{chen2021audio}
H Chen, W Xie, T Afouras, A Nagrani, A Vedaldi, and A Zisserman.
\newblock Audio-visual synchronisation in the wild.
\newblock In {\em Proceedings of the 32nd British Machine Vision Conference}.
  British Machine Vision Association, 2021.

\bibitem{chen2022sound}
Ziyang Chen, David~F Fouhey, and Andrew Owens.
\newblock Sound localization by self-supervised time delay estimation.
\newblock In {\em Computer Vision--ECCV 2022: 17th European Conference, Tel
  Aviv, Israel, October 23--27, 2022, Proceedings, Part XXVI}, pages 489--508.
  Springer, 2022.

\bibitem{chugh2020not}
Komal Chugh, Parul Gupta, Abhinav Dhall, and Ramanathan Subramanian.
\newblock Not made for each other-audio-visual dissonance-based deepfake
  detection and localization.
\newblock In {\em Proceedings of the 28th ACM international conference on
  multimedia}, pages 439--447, 2020.

\bibitem{chung2018voxceleb2}
Joon~Son Chung, Arsha Nagrani, and Andrew Zisserman.
\newblock Voxceleb2: Deep speaker recognition.
\newblock {\em Proc. Interspeech 2018}, pages 1086--1090, 2018.

\bibitem{Chung16}
J.~S. Chung and A. Zisserman.
\newblock Lip reading in the wild.
\newblock In {\em Asian Conference on Computer Vision}, 2016.

\bibitem{chung2016out}
Joon~Son Chung and Andrew Zisserman.
\newblock Out of time: automated lip sync in the wild.
\newblock In {\em Asian conference on computer vision}, pages 251--263.
  Springer, 2016.

\bibitem{chung2019perfect}
Soo-Whan Chung, Joon~Son Chung, and Hong-Goo Kang.
\newblock Perfect match: Improved cross-modal embeddings for audio-visual
  synchronisation.
\newblock In {\em ICASSP 2019-2019 IEEE International Conference on Acoustics,
  Speech and Signal Processing (ICASSP)}, pages 3965--3969. IEEE, 2019.

\bibitem{cozzolino2022audio}
Davide Cozzolino, Matthias Nie{\ss}ner, and Luisa Verdoliva.
\newblock Audio-visual person-of-interest deepfake detection.
\newblock {\em arXiv preprint arXiv:2204.03083}, 2022.

\bibitem{cozzolino2016single}
Davide Cozzolino and Luisa Verdoliva.
\newblock Single-image splicing localization through autoencoder-based anomaly
  detection.
\newblock In {\em 2016 IEEE international workshop on information forensics and
  security (WIFS)}, pages 1--6. IEEE, 2016.

\bibitem{cozzolino2019noiseprint}
Davide Cozzolino and Luisa Verdoliva.
\newblock Noiseprint: a cnn-based camera model fingerprint.
\newblock {\em IEEE Transactions on Information Forensics and Security},
  15:144--159, 2019.

\bibitem{d2017autoencoder}
Dario D'Avino, Davide Cozzolino, Giovanni Poggi, and Luisa Verdoliva.
\newblock Autoencoder with recurrent neural networks for video forgery
  detection.
\newblock {\em arXiv preprint arXiv:1708.08754}, 2017.

\bibitem{deng2009imagenet}
Jia Deng, Wei Dong, Richard Socher, Li-Jia Li, Kai Li, and Li Fei-Fei.
\newblock Imagenet: A large-scale hierarchical image database.
\newblock In {\em 2009 IEEE conference on computer vision and pattern
  recognition}, pages 248--255. Ieee, 2009.

\bibitem{dhamija2018reducing}
Akshay~Raj Dhamija, Manuel G{\"u}nther, and Terrance Boult.
\newblock Reducing network agnostophobia.
\newblock {\em Advances in Neural Information Processing Systems}, 31, 2018.

\bibitem{durall2020watch}
Ricard Durall, Margret Keuper, and Janis Keuper.
\newblock Watch your up-convolution: Cnn based generative deep neural networks
  are failing to reproduce spectral distributions.
\newblock In {\em Proceedings of the IEEE/CVF conference on computer vision and
  pattern recognition}, pages 7890--7899, 2020.

\bibitem{esser2021taming}
Patrick Esser, Robin Rombach, and Bjorn Ommer.
\newblock Taming transformers for high-resolution image synthesis.
\newblock In {\em Proceedings of the IEEE/CVF conference on computer vision and
  pattern recognition}, pages 12873--12883, 2021.

\bibitem{fei2022learning}
Jianwei Fei, Yunshu Dai, Peipeng Yu, Tianrun Shen, Zhihua Xia, and Jian Weng.
\newblock Learning second order local anomaly for general face forgery
  detection.
\newblock In {\em Proceedings of the IEEE/CVF Conference on Computer Vision and
  Pattern Recognition}, pages 20270--20280, 2022.

\bibitem{fox2021videoforensicshq}
Gereon Fox, Wentao Liu, Hyeongwoo Kim, Hans-Peter Seidel, Mohamed Elgharib, and
  Christian Theobalt.
\newblock Videoforensicshq: Detecting high-quality manipulated face videos.
\newblock In {\em 2021 IEEE International Conference on Multimedia and Expo
  (ICME)}, pages 1--6. IEEE, 2021.

\bibitem{frank2020leveraging}
Joel Frank, Thorsten Eisenhofer, Lea Sch{\"o}nherr, Asja Fischer, Dorothea
  Kolossa, and Thorsten Holz.
\newblock Leveraging frequency analysis for deep fake image recognition.
\newblock In {\em International conference on machine learning}, pages
  3247--3258. PMLR, 2020.

\bibitem{gao2021visualvoice}
Ruohan Gao and Kristen Grauman.
\newblock Visualvoice: Audio-visual speech separation with cross-modal
  consistency.
\newblock In {\em 2021 IEEE/CVF Conference on Computer Vision and Pattern
  Recognition (CVPR)}, pages 15490--15500. IEEE, 2021.

\bibitem{NIPS2014_5ca3e9b1}
Ian Goodfellow, Jean Pouget-Abadie, Mehdi Mirza, Bing Xu, David Warde-Farley,
  Sherjil Ozair, Aaron Courville, and Yoshua Bengio.
\newblock Generative adversarial nets.
\newblock In Z. Ghahramani, M. Welling, C. Cortes, N. Lawrence, and K.Q.
  Weinberger, editors, {\em Advances in Neural Information Processing Systems},
  volume~27. Curran Associates, Inc., 2014.

\bibitem{grill2020bootstrap}
Jean-Bastien Grill, Florian Strub, Florent Altch{\'e}, Corentin Tallec, Pierre
  Richemond, Elena Buchatskaya, Carl Doersch, Bernardo Avila~Pires, Zhaohan
  Guo, Mohammad Gheshlaghi~Azar, et~al.
\newblock Bootstrap your own latent-a new approach to self-supervised learning.
\newblock {\em Advances in neural information processing systems},
  33:21271--21284, 2020.

\bibitem{guo2022eyes}
Hui Guo, Shu Hu, Xin Wang, Ming-Ching Chang, and Siwei Lyu.
\newblock Eyes tell all: Irregular pupil shapes reveal gan-generated faces.
\newblock In {\em ICASSP 2022-2022 IEEE International Conference on Acoustics,
  Speech and Signal Processing (ICASSP)}, pages 2904--2908. IEEE, 2022.

\bibitem{haliassos2022leveraging}
Alexandros Haliassos, Rodrigo Mira, Stavros Petridis, and Maja Pantic.
\newblock Leveraging real talking faces via self-supervision for robust forgery
  detection.
\newblock In {\em Proceedings of the IEEE/CVF Conference on Computer Vision and
  Pattern Recognition}, pages 14950--14962, 2022.

\bibitem{haliassos2021lips}
Alexandros Haliassos, Konstantinos Vougioukas, Stavros Petridis, and Maja
  Pantic.
\newblock Lips don't lie: A generalisable and robust approach to face forgery
  detection.
\newblock In {\em Proceedings of the IEEE/CVF conference on computer vision and
  pattern recognition}, pages 5039--5049, 2021.

\bibitem{hara2018can}
Kensho Hara, Hirokatsu Kataoka, and Yutaka Satoh.
\newblock Can spatiotemporal 3d cnns retrace the history of 2d cnns and
  imagenet?
\newblock In {\em Proceedings of the IEEE conference on Computer Vision and
  Pattern Recognition}, pages 6546--6555, 2018.

\bibitem{he2016deep}
Kaiming He, Xiangyu Zhang, Shaoqing Ren, and Jian Sun.
\newblock Deep residual learning for image recognition.
\newblock In {\em Proceedings of the IEEE conference on computer vision and
  pattern recognition}, pages 770--778, 2016.

\bibitem{hendrycks2018deep}
Dan Hendrycks, Mantas Mazeika, and Thomas Dietterich.
\newblock Deep anomaly detection with outlier exposure.
\newblock {\em arXiv preprint arXiv:1812.04606}, 2018.

\bibitem{hu2022mix}
Xixi Hu, Ziyang Chen, and Andrew Owens.
\newblock Mix and localize: Localizing sound sources in mixtures.
\newblock {\em Computer Vision and Pattern Recognition (CVPR)}, 2022.

\bibitem{hu2021dynamic}
Ziheng Hu, Hongtao Xie, Yuxin Wang, Jiahong Li, Zhongyuan Wang, and Yongdong
  Zhang.
\newblock Dynamic inconsistency-aware deepfake video detection.
\newblock In {\em IJCAI}, 2021.

\bibitem{huh2018fighting}
Minyoung Huh, Andrew Liu, Andrew Owens, and Alexei~A Efros.
\newblock Fighting fake news: Image splice detection via learned
  self-consistency.
\newblock {\em European Conference on Computer Vision (ECCV)}, 2018.

\bibitem{jia2018transfer}
Ye Jia, Yu Zhang, Ron Weiss, Quan Wang, Jonathan Shen, Fei Ren, Patrick Nguyen,
  Ruoming Pang, Ignacio Lopez~Moreno, Yonghui Wu, et~al.
\newblock Transfer learning from speaker verification to multispeaker
  text-to-speech synthesis.
\newblock {\em Advances in neural information processing systems}, 31, 2018.

\bibitem{jiang2020deeperforensics}
Liming Jiang, Ren Li, Wayne Wu, Chen Qian, and Chen~Change Loy.
\newblock Deeperforensics-1.0: A large-scale dataset for real-world face
  forgery detection.
\newblock In {\em Proceedings of the IEEE/CVF conference on computer vision and
  pattern recognition}, pages 2889--2898, 2020.

\bibitem{kalyan2018satellite}
Sri Kalyan~Yarlagadda, David G{\"u}era, Paolo Bestagini, Fengqing~Maggie Zhu,
  Stefano Tubaro, and Edward~J Delp.
\newblock Satellite image forgery detection and localization using gan and
  one-class classifier.
\newblock {\em arXiv e-prints}, pages arXiv--1802, 2018.

\bibitem{kay2017kinetics}
Will Kay, Joao Carreira, Karen Simonyan, Brian Zhang, Chloe Hillier, Sudheendra
  Vijayanarasimhan, Fabio Viola, Tim Green, Trevor Back, Paul Natsev, et~al.
\newblock The kinetics human action video dataset.
\newblock {\em arXiv preprint arXiv:1705.06950}, 2017.

\bibitem{NEURIPS_DATASETS_AND_BENCHMARKS2021_d9d4f495}
Hasam Khalid, Shahroz Tariq, Minha Kim, and Simon Woo.
\newblock Fakeavceleb: A novel audio-video multimodal deepfake dataset.
\newblock In J. Vanschoren and S. Yeung, editors, {\em Proceedings of the
  Neural Information Processing Systems Track on Datasets and Benchmarks},
  volume~1, 2021.

\bibitem{khalid2020oc}
Hasam Khalid and Simon~S Woo.
\newblock Oc-fakedect: Classifying deepfakes using one-class variational
  autoencoder.
\newblock In {\em Proceedings of the IEEE/CVF conference on computer vision and
  pattern recognition workshops}, pages 656--657, 2020.

\bibitem{kingma2014adam}
Diederik~P Kingma and Jimmy Ba.
\newblock Adam: A method for stochastic optimization.
\newblock {\em arXiv preprint arXiv:1412.6980}, 2014.

\bibitem{knapp1976generalized}
Charles Knapp and Glifford Carter.
\newblock The generalized correlation method for estimation of time delay.
\newblock {\em IEEE transactions on acoustics, speech, and signal processing},
  1976.

\bibitem{kong2021opengan}
Shu Kong and Deva Ramanan.
\newblock Opengan: Open-set recognition via open data generation.
\newblock In {\em Proceedings of the IEEE/CVF International Conference on
  Computer Vision}, pages 813--822, 2021.

\bibitem{kopuklu2021design}
Okan K{\"o}p{\"u}kl{\"u}, Maja Taseska, and Gerhard Rigoll.
\newblock How to design a three-stage architecture for audio-visual active
  speaker detection in the wild.
\newblock In {\em Proceedings of the IEEE/CVF International Conference on
  Computer Vision}, pages 1193--1203, 2021.

\bibitem{korbar2018cooperative}
Bruno Korbar, Du Tran, and Lorenzo Torresani.
\newblock Cooperative learning of audio and video models from self-supervised
  synchronization.
\newblock {\em Advances in Neural Information Processing Systems}, 31, 2018.

\bibitem{korshunova2017fast}
Iryna Korshunova, Wenzhe Shi, Joni Dambre, and Lucas Theis.
\newblock Fast face-swap using convolutional neural networks.
\newblock In {\em Proceedings of the IEEE international conference on computer
  vision}, pages 3677--3685, 2017.

\bibitem{kr2019towards}
Prajwal KR, Rudrabha Mukhopadhyay, Jerin Philip, Abhishek Jha, Vinay
  Namboodiri, and CV Jawahar.
\newblock Towards automatic face-to-face translation.
\newblock In {\em Proceedings of the 27th ACM international conference on
  multimedia}, pages 1428--1436, 2019.

\bibitem{kwon2021kodf}
Patrick Kwon, Jaeseong You, Gyuhyeon Nam, Sungwoo Park, and Gyeongsu Chae.
\newblock Kodf: A large-scale korean deepfake detection dataset.
\newblock In {\em Proceedings of the IEEE/CVF International Conference on
  Computer Vision}, pages 10744--10753, 2021.

\bibitem{li2020face}
Lingzhi Li, Jianmin Bao, Ting Zhang, Hao Yang, Dong Chen, Fang Wen, and Baining
  Guo.
\newblock Face x-ray for more general face forgery detection.
\newblock In {\em Proceedings of the IEEE/CVF conference on computer vision and
  pattern recognition}, pages 5001--5010, 2020.

\bibitem{li2018exposing}
Yuezun Li and Siwei Lyu.
\newblock Exposing deepfake videos by detecting face warping artifacts.
\newblock {\em arXiv preprint arXiv:1811.00656}, 2018.

\bibitem{liang2017enhancing}
Shiyu Liang, Yixuan Li, and Rayadurgam Srikant.
\newblock Enhancing the reliability of out-of-distribution image detection in
  neural networks.
\newblock {\em arXiv preprint arXiv:1706.02690}, 2017.

\bibitem{liu2018generating}
Peter~J Liu, Mohammad Saleh, Etienne Pot, Ben Goodrich, Ryan Sepassi, Lukasz
  Kaiser, and Noam Shazeer.
\newblock Generating wikipedia by summarizing long sequences.
\newblock In {\em International Conference on Learning Representations}, 2018.

\bibitem{liu2019generative}
Yezheng Liu, Zhe Li, Chong Zhou, Yuanchun Jiang, Jianshan Sun, Meng Wang, and
  Xiangnan He.
\newblock Generative adversarial active learning for unsupervised outlier
  detection.
\newblock {\em IEEE Transactions on Knowledge and Data Engineering},
  32(8):1517--1528, 2019.

\bibitem{liu2020global}
Zhengzhe Liu, Xiaojuan Qi, and Philip~HS Torr.
\newblock Global texture enhancement for fake face detection in the wild.
\newblock In {\em Proceedings of the IEEE/CVF conference on computer vision and
  pattern recognition}, pages 8060--8069, 2020.

\bibitem{loshchilov2016sgdr}
Ilya Loshchilov and Frank Hutter.
\newblock Sgdr: Stochastic gradient descent with warm restarts.
\newblock {\em arXiv preprint arXiv:1608.03983}, 2016.

\bibitem{ma2021towards}
Pingchuan Ma, Brais Martinez, Stavros Petridis, and Maja Pantic.
\newblock Towards practical lipreading with distilled and efficient models.
\newblock In {\em ICASSP 2021-2021 IEEE International Conference on Acoustics,
  Speech and Signal Processing (ICASSP)}, pages 7608--7612. IEEE, 2021.

\bibitem{majumder2021move2hear}
Sagnik Majumder, Ziad Al-Halah, and Kristen Grauman.
\newblock Move2hear: Active audio-visual source separation.
\newblock In {\em Proceedings of the IEEE/CVF International Conference on
  Computer Vision}, pages 275--285, 2021.

\bibitem{malik2010audio}
Hafiz Malik and Hany Farid.
\newblock Audio forensics from acoustic reverberation.
\newblock In {\em 2010 IEEE International Conference on Acoustics, Speech and
  Signal Processing}, 2010.

\bibitem{martinez2020lipreading}
Brais Martinez, Pingchuan Ma, Stavros Petridis, and Maja Pantic.
\newblock Lipreading using temporal convolutional networks.
\newblock In {\em ICASSP 2020-2020 IEEE International Conference on Acoustics,
  Speech and Signal Processing (ICASSP)}, pages 6319--6323. IEEE, 2020.

\bibitem{mittal2020emotions}
Trisha Mittal, Uttaran Bhattacharya, Rohan Chandra, Aniket Bera, and Dinesh
  Manocha.
\newblock Emotions don't lie: An audio-visual deepfake detection method using
  affective cues.
\newblock In {\em Proceedings of the 28th ACM international conference on
  multimedia}, pages 2823--2832, 2020.

\bibitem{mo2022_ezvsl}
Shentong Mo and Pedro Morgado.
\newblock Localizing visual sounds the easy way.
\newblock In {\em European Conference on Computer Vision (ECCV)}, 2022.

\bibitem{morgado2021audio}
Pedro Morgado, Nuno Vasconcelos, and Ishan Misra.
\newblock Audio-visual instance discrimination with cross-modal agreement.
\newblock In {\em Proceedings of the IEEE/CVF Conference on Computer Vision and
  Pattern Recognition}, pages 12475--12486, 2021.

\bibitem{nirkin2019fsgan}
Yuval Nirkin, Yosi Keller, and Tal Hassner.
\newblock Fsgan: Subject agnostic face swapping and reenactment.
\newblock In {\em Proceedings of the IEEE/CVF international conference on
  computer vision}, pages 7184--7193, 2019.

\bibitem{oord2018representation}
Aaron van~den Oord, Yazhe Li, and Oriol Vinyals.
\newblock Representation learning with contrastive predictive coding.
\newblock {\em arXiv preprint arXiv:1807.03748}, 2018.

\bibitem{owens2018audio}
Andrew Owens and Alexei~A Efros.
\newblock Audio-visual scene analysis with self-supervised multisensory
  features.
\newblock {\em European Conference on Computer Vision (ECCV)}, 2018.

\bibitem{perez2019deep}
Daniel P{\'e}rez-Cabo, David Jim{\'e}nez-Cabello, Artur Costa-Pazo, and
  Roberto~J L{\'o}pez-Sastre.
\newblock Deep anomaly detection for generalized face anti-spoofing.
\newblock In {\em Proceedings of the IEEE/CVF Conference on Computer Vision and
  Pattern Recognition Workshops}, pages 0--0, 2019.

\bibitem{perov2020deepfacelab}
Ivan Perov, Daiheng Gao, Nikolay Chervoniy, Kunlin Liu, Sugasa Marangonda,
  Chris Um{\'e}, Mr Dpfks, Carl~Shift Facenheim, RP Luis, Jian Jiang, et~al.
\newblock Deepfacelab: A simple, flexible and extensible face swapping
  framework.
\newblock 2020.

\bibitem{pidhorskyi2018generative}
Stanislav Pidhorskyi, Ranya Almohsen, and Gianfranco Doretto.
\newblock Generative probabilistic novelty detection with adversarial
  autoencoders.
\newblock {\em Advances in neural information processing systems}, 31, 2018.

\bibitem{prajwal2020lip}
KR Prajwal, Rudrabha Mukhopadhyay, Vinay~P Namboodiri, and CV Jawahar.
\newblock A lip sync expert is all you need for speech to lip generation in the
  wild.
\newblock In {\em Proceedings of the 28th ACM International Conference on
  Multimedia}, pages 484--492, 2020.

\bibitem{qian2020thinking}
Yuyang Qian, Guojun Yin, Lu Sheng, Zixuan Chen, and Jing Shao.
\newblock Thinking in frequency: Face forgery detection by mining
  frequency-aware clues.
\newblock In {\em European conference on computer vision}, pages 86--103.
  Springer, 2020.

\bibitem{radford2019language}
Alec Radford, Jeffrey Wu, Rewon Child, David Luan, Dario Amodei, Ilya
  Sutskever, et~al.
\newblock Language models are unsupervised multitask learners.
\newblock {\em OpenAI blog}, 1(8):9, 2019.

\bibitem{ramesh2022hierarchical}
Aditya Ramesh, Prafulla Dhariwal, Alex Nichol, Casey Chu, and Mark Chen.
\newblock Hierarchical text-conditional image generation with clip latents.
\newblock {\em arXiv preprint arXiv:2204.06125}, 2022.

\bibitem{razavi2019generating}
Ali Razavi, Aaron Van~den Oord, and Oriol Vinyals.
\newblock Generating diverse high-fidelity images with vq-vae-2.
\newblock {\em Advances in neural information processing systems}, 32, 2019.

\bibitem{rossler2019faceforensics++}
Andreas Rossler, Davide Cozzolino, Luisa Verdoliva, Christian Riess, Justus
  Thies, and Matthias Nie{\ss}ner.
\newblock Faceforensics++: Learning to detect manipulated facial images.
\newblock In {\em Proceedings of the IEEE/CVF international conference on
  computer vision}, pages 1--11, 2019.

\bibitem{ruff2019deep}
Lukas Ruff, Robert~A Vandermeulen, Nico G{\"o}rnitz, Alexander Binder, Emmanuel
  M{\"u}ller, Klaus-Robert M{\"u}ller, and Marius Kloft.
\newblock Deep semi-supervised anomaly detection.
\newblock {\em arXiv preprint arXiv:1906.02694}, 2019.

\bibitem{sabokrou2018adversarially}
Mohammad Sabokrou, Mohammad Khalooei, Mahmood Fathy, and Ehsan Adeli.
\newblock Adversarially learned one-class classifier for novelty detection.
\newblock In {\em Proceedings of the IEEE conference on computer vision and
  pattern recognition}, pages 3379--3388, 2018.

\bibitem{salimans2017pixelcnn++}
Tim Salimans, Andrej Karpathy, Xi Chen, and Diederik~P Kingma.
\newblock Pixelcnn++: Improving the pixelcnn with discretized logistic mixture
  likelihood and other modifications.
\newblock {\em arXiv preprint arXiv:1701.05517}, 2017.

\bibitem{schlegl2017unsupervised}
Thomas Schlegl, Philipp Seeb{\"o}ck, Sebastian~M Waldstein, Ursula
  Schmidt-Erfurth, and Georg Langs.
\newblock Unsupervised anomaly detection with generative adversarial networks
  to guide marker discovery.
\newblock In {\em International conference on information processing in medical
  imaging}, pages 146--157. Springer, 2017.

\bibitem{siarohin2019first}
Aliaksandr Siarohin, St{\'e}phane Lathuili{\`e}re, Sergey Tulyakov, Elisa
  Ricci, and Nicu Sebe.
\newblock First order motion model for image animation.
\newblock {\em Advances in Neural Information Processing Systems}, 32, 2019.

\bibitem{song2017pixeldefend}
Yang Song, Taesup Kim, Sebastian Nowozin, Stefano Ermon, and Nate Kushman.
\newblock Pixeldefend: Leveraging generative models to understand and defend
  against adversarial examples.
\newblock {\em arXiv preprint arXiv:1710.10766}, 2017.

\bibitem{srivastava2014dropout}
Nitish Srivastava, Geoffrey Hinton, Alex Krizhevsky, Ilya Sutskever, and Ruslan
  Salakhutdinov.
\newblock Dropout: a simple way to prevent neural networks from overfitting.
\newblock {\em The journal of machine learning research}, 15(1):1929--1958,
  2014.

\bibitem{tao2021someone}
Ruijie Tao, Zexu Pan, Rohan~Kumar Das, Xinyuan Qian, Mike~Zheng Shou, and
  Haizhou Li.
\newblock Is someone speaking? exploring long-term temporal features for
  audio-visual active speaker detection.
\newblock In {\em Proceedings of the 29th ACM International Conference on
  Multimedia}, pages 3927--3935, 2021.

\bibitem{van2016conditional}
Aaron Van~den Oord, Nal Kalchbrenner, Lasse Espeholt, Oriol Vinyals, Alex
  Graves, et~al.
\newblock Conditional image generation with pixelcnn decoders.
\newblock {\em Advances in neural information processing systems}, 29, 2016.

\bibitem{van2017neural}
Aaron Van Den~Oord, Oriol Vinyals, et~al.
\newblock Neural discrete representation learning.
\newblock {\em Advances in neural information processing systems}, 30, 2017.

\bibitem{van2016pixel}
Aaron Van~Oord, Nal Kalchbrenner, and Koray Kavukcuoglu.
\newblock Pixel recurrent neural networks.
\newblock In {\em International conference on machine learning}, pages
  1747--1756. PMLR, 2016.

\bibitem{vaswani2017attention}
Ashish Vaswani, Noam Shazeer, Niki Parmar, Jakob Uszkoreit, Llion Jones,
  Aidan~N Gomez, {\L}ukasz Kaiser, and Illia Polosukhin.
\newblock Attention is all you need.
\newblock {\em Advances in neural information processing systems}, 30, 2017.

\bibitem{wang2019cnn}
Sheng-Yu Wang, Oliver Wang, Richard Zhang, Andrew Owens, and Alexei~A Efros.
\newblock Cnn-generated images are surprisingly easy to spot... for now.
\newblock {\em Computer Vision and Pattern Recognition (CVPR)}, 2020.

\bibitem{wu2019mantra}
Yue Wu, Wael AbdAlmageed, and Premkumar Natarajan.
\newblock Mantra-net: Manipulation tracing network for detection and
  localization of image forgeries with anomalous features.
\newblock In {\em Proceedings of the IEEE/CVF Conference on Computer Vision and
  Pattern Recognition}, pages 9543--9552, 2019.

\bibitem{yi2020audio}
Ran Yi, Zipeng Ye, Juyong Zhang, Hujun Bao, and Yong-Jin Liu.
\newblock Audio-driven talking face video generation with learning-based
  personalized head pose.
\newblock {\em arXiv preprint arXiv:2002.10137}, 2020.

\bibitem{yu2022scaling}
Jiahui Yu, Yuanzhong Xu, Jing~Yu Koh, Thang Luong, Gunjan Baid, Zirui Wang,
  Vijay Vasudevan, Alexander Ku, Yinfei Yang, Burcu~Karagol Ayan, et~al.
\newblock Scaling autoregressive models for content-rich text-to-image
  generation.
\newblock {\em arXiv preprint arXiv:2206.10789}, 2022.

\bibitem{zenati2018adversarially}
Houssam Zenati, Manon Romain, Chuan-Sheng Foo, Bruno Lecouat, and Vijay
  Chandrasekhar.
\newblock Adversarially learned anomaly detection.
\newblock In {\em 2018 IEEE International conference on data mining (ICDM)},
  pages 727--736. IEEE, 2018.

\bibitem{zeng2021contrastive}
Zhaoyang Zeng, Daniel McDuff, Yale Song, et~al.
\newblock Contrastive learning of global and local video representations.
\newblock {\em Advances in Neural Information Processing Systems},
  34:7025--7040, 2021.

\bibitem{zhang2020hybrid}
Hongjie Zhang, Ang Li, Jie Guo, and Yanwen Guo.
\newblock Hybrid models for open set recognition.
\newblock In {\em European Conference on Computer Vision}, pages 102--117.
  Springer, 2020.

\bibitem{zhang2017s3fd}
Shifeng Zhang, Xiangyu Zhu, Zhen Lei, Hailin Shi, Xiaobo Wang, and Stan~Z Li.
\newblock S3fd: Single shot scale-invariant face detector.
\newblock In {\em Proceedings of the IEEE international conference on computer
  vision}, pages 192--201, 2017.

\bibitem{zhao2021learning}
Tianchen Zhao, Xiang Xu, Mingze Xu, Hui Ding, Yuanjun Xiong, and Wei Xia.
\newblock Learning self-consistency for deepfake detection.
\newblock In {\em Proceedings of the IEEE/CVF international conference on
  computer vision}, pages 15023--15033, 2021.

\bibitem{zheng2021exploring}
Yinglin Zheng, Jianmin Bao, Dong Chen, Ming Zeng, and Fang Wen.
\newblock Exploring temporal coherence for more general video face forgery
  detection.
\newblock In {\em Proceedings of the IEEE/CVF International Conference on
  Computer Vision}, pages 15044--15054, 2021.

\bibitem{zhou2020sep}
Hang Zhou, Xudong Xu, Dahua Lin, Xiaogang Wang, and Ziwei Liu.
\newblock Sep-stereo: Visually guided stereophonic audio generation by
  associating source separation.
\newblock In {\em Proceedings of the European Conference on Computer Vision
  (ECCV)}, 2020.

\bibitem{zhou2021face}
Tianfei Zhou, Wenguan Wang, Zhiyuan Liang, and Jianbing Shen.
\newblock Face forensics in the wild.
\newblock In {\em Proceedings of the IEEE/CVF conference on computer vision and
  pattern recognition}, pages 5778--5788, 2021.

\bibitem{zhou2021joint}
Yipin Zhou and Ser-Nam Lim.
\newblock Joint audio-visual deepfake detection.
\newblock In {\em Proceedings of the IEEE/CVF International Conference on
  Computer Vision}, pages 14800--14809, 2021.

\bibitem{zong2018deep}
Bo Zong, Qi Song, Martin~Renqiang Min, Wei Cheng, Cristian Lumezanu, Daeki Cho,
  and Haifeng Chen.
\newblock Deep autoencoding gaussian mixture model for unsupervised anomaly
  detection.
\newblock In {\em International conference on learning representations}, 2018.

\end{thebibliography}
}

\clearpage
\appendix
\supparxiv{
\setcounter{page}{1}
\twocolumn[{%
\renewcommand\twocolumn[1][]{#1}%
\begin{center}
    \vspace{-2.0em}
    {\bf \large Supplementary Material:\\Self-Supervised Video Forensics by Audio-Visual Anomaly Detection}
    \vspace{2.0em}
\end{center}

}]

}{}

\renewcommand{\thesection}{A.\arabic{section}}
\setcounter{section}{0}

\section{Video Results}
We provide some  qualitative video results of some random samples from the FakeAVCeleb dataset~\cite{NEURIPS_DATASETS_AND_BENCHMARKS2021_d9d4f495} in \href{https://cfeng16.github.io/audio-visual-forensics/}{our webpage} with audio. We show ``ground truth'' outputs from the synchronization model and autoregressive predictions over time. We also show a score indicating the probability that the example is fake~(\cref{eq:loss}, main paper). We use the time delay distribution as our feature set, and display the predictions as a heat map. Each step $\bx_t$ of the predicted outputs was obtained by providing the ground truth features from times $\bx_1, \bx_2, ..., \bx_{t-1}$ into the autoregressive model.

\section{Cross-dataset Generalization}
\label{appendix:crossdataset}
We also experiment with the another audio-driven video editing method LipGAN~\cite{kr2019towards}. We test on a test set consisting of 100 real videos and 100 fake videos, which are based on the real videos of test set from FakeAVCeleb~\cite{NEURIPS_DATASETS_AND_BENCHMARKS2021_d9d4f495}. We show results on \cref{appendix-tab:crossdataset}. Our method leverages continuous time delay distribution as in \cref{feature_analysis} and outperforms many supervised methods although it is only trained on real videos, indicating that our method possesses certain generalization capability to different manipulation methods. 

\begin{table}[h!]
\centering
\upvspacefig
\resizebox{1\columnwidth}{!}{
\begin{tabular}{llccc}
\toprule
\multicolumn{2}{c}{\multirow{2}{*}{Method}} & \multirow{2}{*}{Modality} &\multicolumn{2}{c}{LipGAN~\cite{kr2019towards}}\\
\cmidrule(lr){4-5}
\multicolumn{2}{c}{}           &             & AP          & AUC         \\
\midrule
\multirow{5}{*}{\shortstack[c]{Supervised \\ (transfer)}}
 & Xception~\cite{rossler2019faceforensics++}   %
   & $\mathcal{V}$  & 67.6   & 65.5  \\
                            & LipForensics~\cite{haliassos2021lips}   &  $\mathcal{V}$ &   82.2    &   85.2      \\
                            & AD DFD~\cite{zhou2021joint}   & $\mathcal{AV}$      &  82.7     &   84.8   \\
                            & FTCN~\cite{zheng2021exploring} & $\mathcal{V}$           &    76.9      &  73.9         \\
                            & RealForensics~\cite{haliassos2022leveraging}   &        $\mathcal{V}$ &  \textbf{94.3}    & \textbf{96.5}          \\
\midrule
\multirow{3}{*}{Unsupervised}                  
& AVBYOL~\cite{haliassos2022leveraging, grill2020bootstrap} &$\mathcal{AV}$   &  65.0    & 73.0 \\

& VQ-GAN ~\cite{esser2021taming} & $\mathcal{V}$   &  50.7     &  50.4     \\

\cdashlinelr{2-5}
& Ours     & $\mathcal{AV}$    &     \textbf{93.3}      &   \textbf{94.1}          \\

\bottomrule
\end{tabular}
 }
\caption{\textbf{Generalization to LipGAN method}~\cite{kr2019towards}. AP scores ($\%$) and AUC scores ($\%$) are reported on real videos of test set from FakeAVCeleb~\cite{NEURIPS_DATASETS_AND_BENCHMARKS2021_d9d4f495}. Fake videos are manipulated by LipGAN~\cite{kr2019towards}. Supervised methods are trained on FakeAVCeleb dataset~\cite{NEURIPS_DATASETS_AND_BENCHMARKS2021_d9d4f495}. Ours is trained on LRS2~\cite{Afouras18c} and uses the feature set of continuous time delay distribution. Best results are in \textbf{bold}.}
\label{appendix-tab:crossdataset}
\end{table}

\section{Ablation Study}
\begin{figure}[t]
    \centering
    \upvspacefig
    \includegraphics[width=0.95\linewidth]{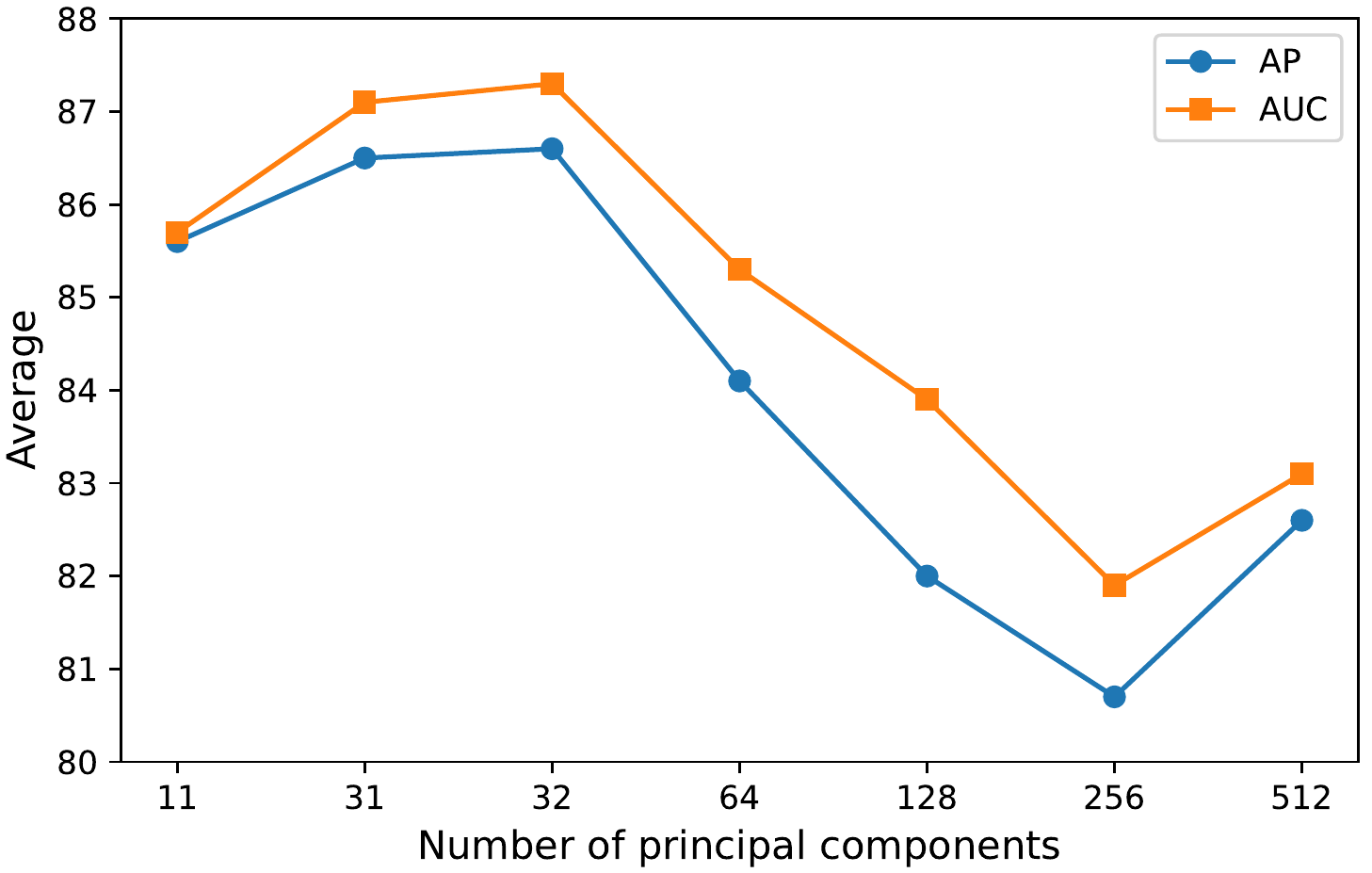}
    \caption{\textbf{Number of PCA projections.} We evaluate with different number of principal components for the model that obtains a feature set by projecting the audio-visual feature activations to a lower dimensional space using PCA. We report average AP scores ($\%$) and AUC scores ($\%$) on FakeAVCeleb dataset~\cite{NEURIPS_DATASETS_AND_BENCHMARKS2021_d9d4f495}.} 
    \label{number_pca}
\end{figure}

\paragraph{Feature activation. }
We study the influence of the number of principal components $D$ for the variation of the anomaly detection model that projects the audio-visual feature activations into a lower dimensional space. We set $D$ to $\{11, 31, 32, 64, 128, 256, 512\}$ and keep other hyperparameters the same. \cref{number_pca} shows that as the $D$ increases, the accuracy of the anomaly detection model decreases. The reason might be that the higher dimensional prediction problem is also more challenging.

\section{Feature Set Variations}
\label{appendix:feature_set}
We describe more details about building the autoregressive model on some of our feature sets in this section.

\mypar{Binary cross entropy (BCE) model.} 
We assume that the $2\tau+1$ possible time delays of each time step are independent, and use BCE loss for probability $S(i,j)$ of each time delay.  
Thus, we can decompose $ p_{\theta}(\mathbf{x}_{1}, \mathbf{x}_{2}, \cdots, \mathbf{x}_{N})$ as in \cref{prob_decom}:
\begin{equation}\label{prob_decom}
p_{\theta}(\mathbf{x}_{1}, \mathbf{x}_{2}, \cdots , \mathbf{x}_{N}) = 
\prod_{i=0}^{N-1}\prod_{q=1}^{2\tau + 1}p_{\theta}\left(x_{i+1, q}|\mathbf{x}_{1},
\cdots,\mathbf{x}_{i}\right),
\end{equation}
We then maximize $p_{\theta}\left(x_{i+1, q}|\mathbf{x}_{1},\cdots,\mathbf{x}_{i}\right)$ using BCE loss:
\begin{equation}\label{bce}
    L(\hat \bx_i, \bx_i) = - \sum_{j=1}^{2\tau+1} x_{i,j} \log(\hat x_{i,j}) + (1-x_{i,j})\log(1 - \hat{x}_{i,j}).
\end{equation}

We constrain the prediction $\hat{x}_{i,j}$ to the range of $[0, 1]$ via a sigmoid function.
\begin{figure*}[t!]
    \centering
    \upvspacefig
 \includegraphics[width=1\linewidth]{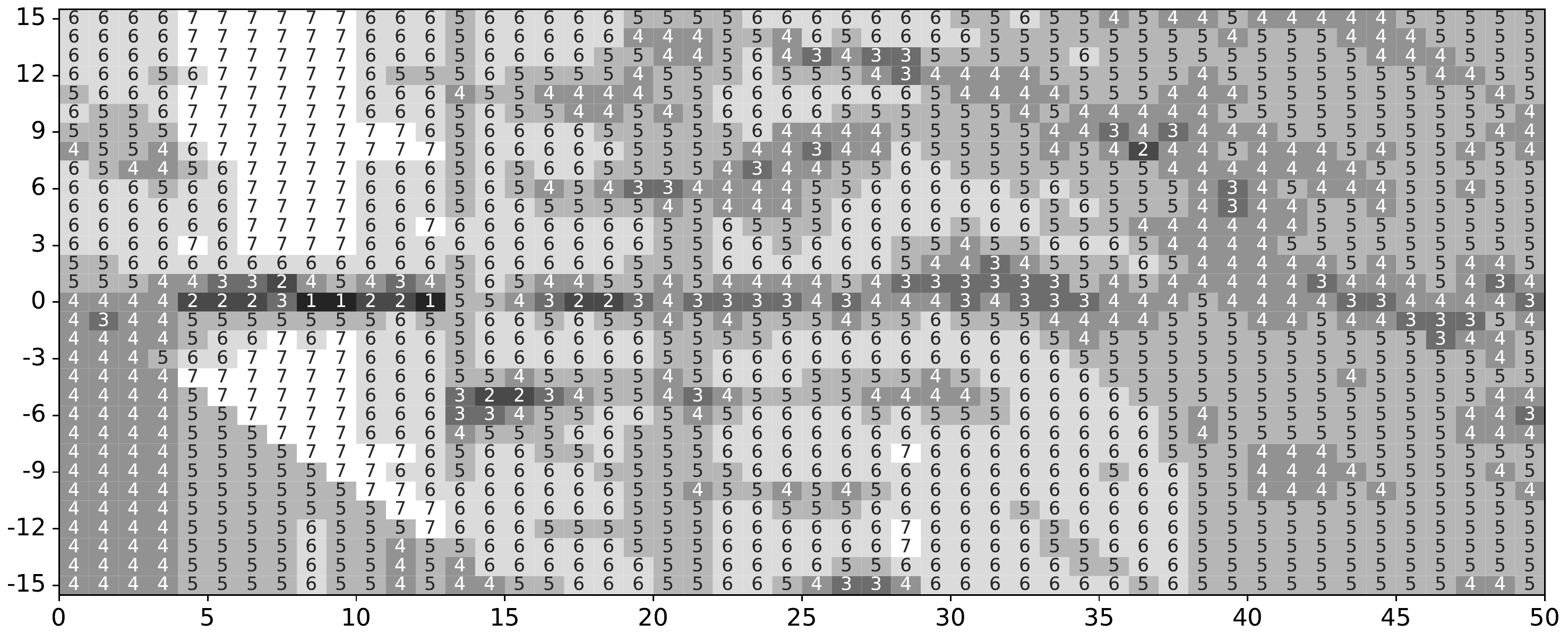}
    \caption{\textbf{Visualization of discrete probability grid.} We use K-means clustering to quantize the probability space to convert continuous synchronization probability $S(i,j)$ to discrete probability bin $\hat{S}(i,j)$. Then we build a autoregressive Transformer~\cite{vaswani2017attention} model on the probability grid. }
    \label{pixelcnn}
\end{figure*}

\mypar{Discrete 2D probability model.} 
The {\em discrete prob.} model generates the entire 2D frame/time delay matrix autoregressively in ``raster scan" order, similar to models such as PixelCNN~\cite{van2016pixel,salimans2017pixelcnn++}. We unroll the time delay distribution sequence into 2D grid like a grayscale image as shown in \cref{pixelcnn}. We use $K$-means ($K=8$) clustering to quantize each entry and assume each probability $\hat{S}(i,j)$ of time delay not only depends on the previous time delay distributions but also the previous probabilities within the same distribution. Then we decompose $ p_{\theta}(\mathbf{x}_{1}, \mathbf{x}_{2}, \cdots , \mathbf{x}_{N})$:
\begin{equation}\label{eq_pixelcnn}
\begin{split}
    p_{\theta}&(\mathbf{x}_{1}, \mathbf{x}_{2}, \cdots , \mathbf{x}_{N}) = 
    \prod_{i=0}^{N-1}\prod_{q=1}^{2\tau + 1}\\
    &p_{\theta}(x_{i+1, q}|\mathbf{x}_{1},
    \cdots,\mathbf{x}_{i};x_{i+1, 1},\cdots,x_{i+1,q-1}).
\end{split}
\end{equation}
We utilize a similar loss function in PixelCNN~\cite{salimans2017pixelcnn++,van2016pixel} to supervise the autoregressive model.

\section{Implementation Details}
\label{appendix:imple}
\paragraph{Audio-visual synchronization model.}
Following prior work~\cite{Afouras20b,chen2021audio,korbar2018cooperative}, we utilize curriculum learning to train the audio-visual synchronization model. Specifically, we use the two-stage training procedure. During the first phase, the negatives are from different videos, while for the second stage, the negatives are sampled within the same videos randomly (the starting time steps are sampled randomly).

\mypar{Anomaly detection model.}
We process each input vector $\bx_i$ with an affine transformation, before passing it into the autoregressive model. This projects the input (e.g., a time delay distribution $S_i \in \mathbb{R}^{31}$) to $\mathbb{R}^{256}$. Also, we add an affine transformation to project the embedding back into the feature set space, e.g., $\mathbb{R}^{256} \rightarrow  \mathbb{R}^{31}$ for the time delay distribution. We use learnable positional encodings for both the synchronization and autoregressive models.

\mypar{Hyperparameters.}
For the synchronization model, we use Adam~\cite{kingma2014adam} with a learning rate of  $1\times10^{-4}$, with a batch size of 16 for the first phase and 40 for the second. During the second stage, we sample four 5-frame short clips per video. For the autoregressive model (anomaly detection model), we use the Adam optimizer~\cite{kingma2014adam} with $1\times10^{-3}$ learning rate, weight decay of $1\times10^{-6}$ and warm-up and cosine learning rate decay~\cite{loshchilov2016sgdr} strategies. We use a batch size of 16 and the dropout~\cite{srivastava2014dropout} rate of $0.1$. We train the synchronization model on 8 NVIDIA A40 GPUs and use a single GeForce RTX 2080 Ti GPU for the autoregressive model.

\begin{figure}[t!]
    \centering
    \includegraphics[width=0.95\linewidth]{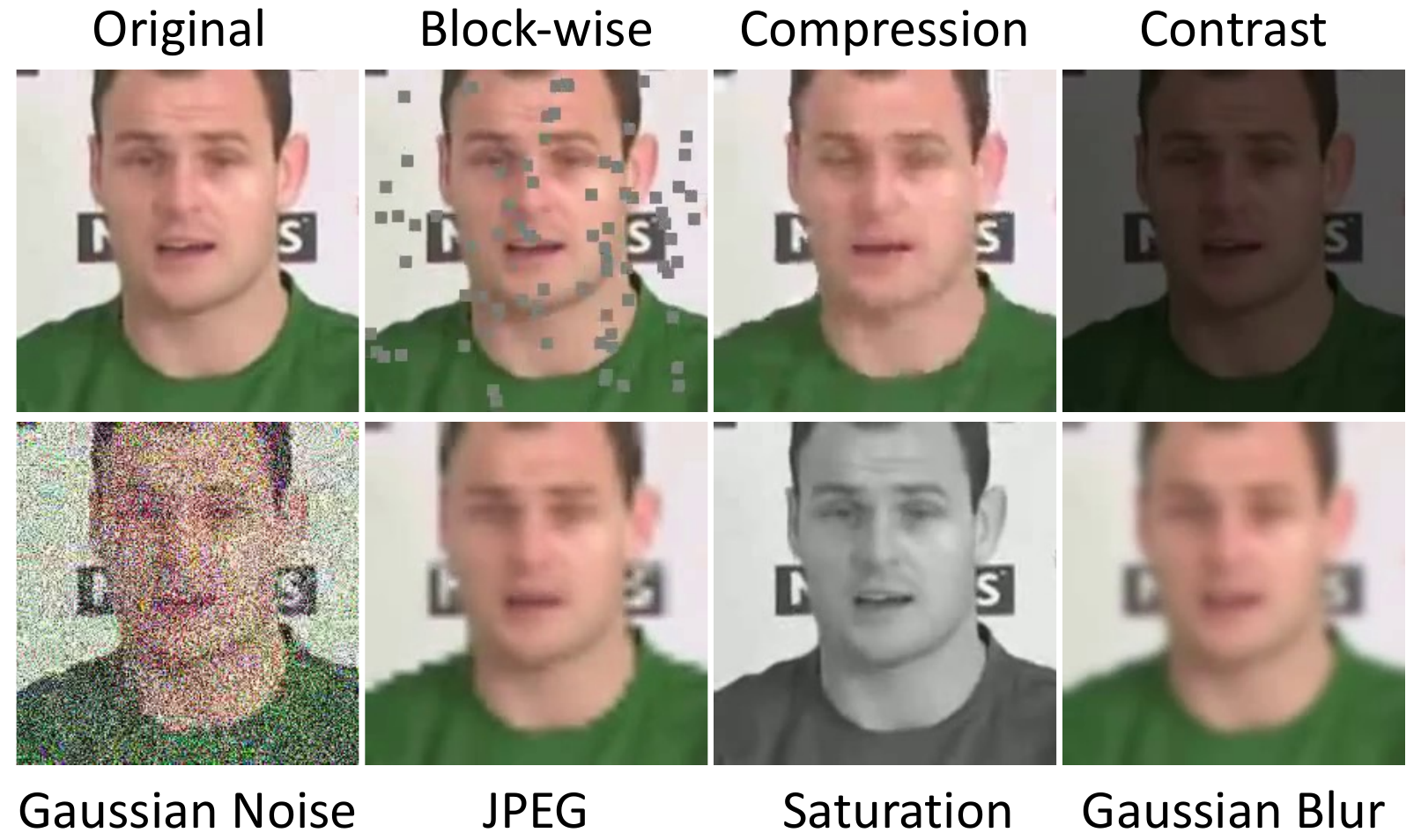}
    \caption{\textbf{Visualization of corrupted images.} Examples of the corruptions taken into account at intensity level 5. The set of corruptions is introduced in Jiang et al.~\cite{jiang2020deeperforensics}, and it consists of color saturation, local block-wise distortion, color contrast, Gaussian blur, white Gaussian noise, JPEG compression, and video compression rate change.}
    \label{corrupted_images}
\end{figure}
\section{Visualization of perturbed images}
We visualize some images used for unseen perturbations robustness test in \cref{corrupted_images}.

\section{Temporal localization}

\begin{figure}[h]
    \centering

    \includegraphics[width=\linewidth]{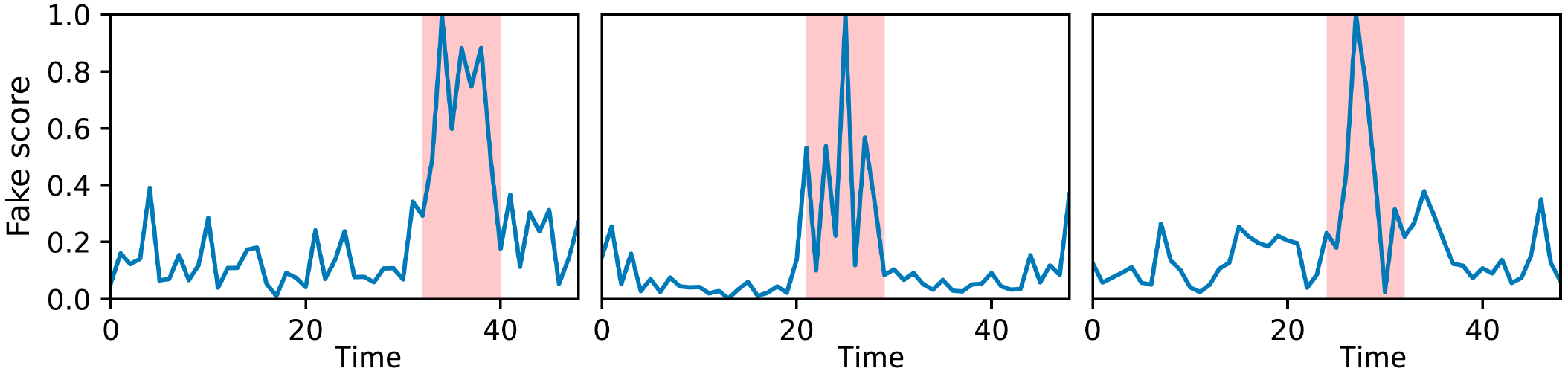}

        \caption{\textbf{Temporal localization.} We manipulate a random 9-time step interval~(\textcolor{pink}{pink region}). We show the fake score~(negative log-likelihood) for each frame. We scale the $y$ axis by the maximum score. }

    \label{temporal_localization_arxiv}
\end{figure}

To test the temporal localization ability of our method, we selected random 5-frame subsequences from real videos of FakeAVCeleb~\cite{NEURIPS_DATASETS_AND_BENCHMARKS2021_d9d4f495} test set and manipulated frames using Wav2Lip~\cite{prajwal2020lip} with audio (since none of our benchmark datasets provided relevant videos for the task). Our model predicts the fake score for each frame~(\cref{temporal_localization_arxiv}). We can see that the score was significantly higher for manipulated frames, suggesting that our model is able to temporally localize manipulations. Besides, we can achieve top-5 accuracy of 92.0\%. It is worth mentioning that all supervised and unsupervised baselines can not localize the manipulated frames temporally. 

\section{Robustness to background noise}

\begin{figure}[h]
    \centering
    \upvspacefig
    \includegraphics[width=0.45\linewidth]{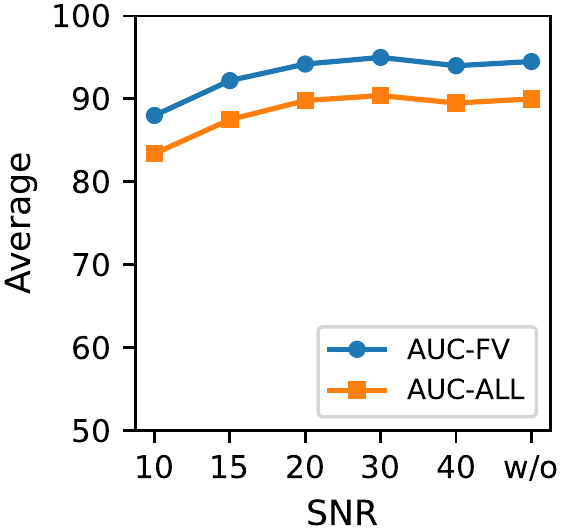}
    \caption{\textbf{Robustness to background noise.} AUC scores $(\%)$ of our detector as a function of signal-to-noise ratio (SNR) when Gaussian noise is added to the audio stream. ``w/o" means no noise is added. AVG-ALL means the average over all the categories. AVG-FV represents the average over four fake video categories.} 
    \label{noise}
\end{figure}

We test our method when the audio stream of FakeAVCeleb~\cite{NEURIPS_DATASETS_AND_BENCHMARKS2021_d9d4f495} test set is added Gaussian noise under different signal-to-noise ratio (SNR). As shown in \cref{noise}, our detector can basically maintain its performance when SNR decreases, indicating that our method is robust to the background Gaussian noise (our method has never seen audio Gaussian noise during training). 

\end{document}